%% file: main.tex
\documentclass[sigconf]{acmart}
\AtBeginDocument{%
  }

\usepackage{graphicx}
\usepackage[utf8]{inputenc}
\usepackage{csquotes}
\usepackage{color}
\usepackage[normalem]{ulem}
\usepackage{tabularray}
\usepackage{xcolor} 
\usepackage{tcolorbox} 
\usepackage{array}
\usepackage{hhline}
\usepackage{arydshln}
\usepackage{amsfonts}
\usepackage{mathtools}
\usepackage{subcaption}
\usepackage{setspace}
\usepackage{booktabs}
\usepackage{makecell}
\usepackage{listings}
\usepackage{pgf-pie} 
\usepackage{pgfplots} 
\usepackage{amsmath}
\usepackage{amsfonts}
\usepackage{enumitem}
\usepackage{hyperref}
\usepackage{xcolor}
\usepackage{listings}
\usepackage{tikz}
\usepackage{pgf-pie}
\usepackage{caption}
\usepackage{subcaption}
\usepackage{booktabs}
\usepackage{array}
\usepackage{graphicx}
\usepackage{pifont}

\definecolor{mygreen}{RGB}{34,178,34}
\definecolor{myred}{RGB}{178,34,34}
\newcommand{\cmark}{{\color{mygreen} \ding{51}}}
\newcommand{\xmark}{{\color{myred} \ding{55}}}

\newcommand{\default}[1]{%
    \tcbox[
        size=small,
        boxrule=0.5pt,
        arc=3pt,
        colback=gray!20,
        colframe=gray!40,
        fontupper=\small\sffamily,
        nobeforeafter,
        tcbox raise base
    ]{#1}
}

\newcommand{\usa}[1]{%
    \tcbox[
        size=small,
        boxrule=0.5pt,
        arc=3pt,
        colback=blue!10,
        colframe=blue!40,
        fontupper=\small\sffamily,
        nobeforeafter,
        tcbox raise base
    ]{#1}
}

\newcommand{\chn}[1]{%
    \tcbox[
        size=small,
        boxrule=0.5pt,
        arc=3pt,
        colback=red!10,
        colframe=red!40,
        fontupper=\small\sffamily,
        nobeforeafter,
        tcbox raise base
    ]{#1}
}

\copyrightyear{2026}
\acmYear{2026}
\setcopyright{cc}
\setcctype{by}
\acmConference[KDD '26]{Proceedings of the 32nd ACM SIGKDD Conference on Knowledge Discovery and Data Mining V.1}{August 09--13, 2026}{Jeju Island, Republic of Korea}
\acmBooktitle{Proceedings of the 32nd ACM SIGKDD Conference on Knowledge Discovery and Data Mining V.1 (KDD '26), August 09--13, 2026, Jeju Island, Republic of Korea}
\acmPrice{}
\acmDOI{10.1145/3770854.3785691}
\acmISBN{979-8-4007-2258-5/2026/08}

\begin{document}

\title{\textsc{MemeBridge}: A Dataset for Benchmarking and Mitigating the Bidirectional Cultural Gap in Meme Interpretation}

\author{Hangxiao Zhu}
\email{hangxiao@tamu.edu}
\orcid{0009-0007-1394-3410}
\affiliation{%
  \institution{Texas A\&M University}
  \city{College Station}
  \country{USA}
}

\author{Suliu Qin}
\email{Sophie508727@gmail.com}
\orcid{0009-0004-0131-2783}
\affiliation{%
  \institution{Independent Researcher}
  \city{College Station}
  \country{USA}
}

\author{Zhuoyan Li}
\email{li4178@purdue.edu}
\orcid{0000-0001-7143-9262}
\affiliation{%
  \institution{Purdue University}
  \city{West Lafayette}
  \country{USA}
}

\author{Ming Jiang}
\email{ming.jiang@wisc.edu}
\orcid{0000-0002-3604-166X}
\affiliation{%
 \institution{University of Wisconsin–Madison}
 \city{Madison}
 \country{USA}
}

\author{Yu Zhang}
\email{yuzhang@tamu.edu}
\orcid{0000-0003-0540-6758}
\affiliation{%
  \institution{Texas A\&M University}
  \city{College Station}
  \country{USA}
}

\author{Meng Xia}
\email{mengxia@tamu.edu}
\orcid{0000-0002-2676-9032}
\affiliation{%
  \institution{Texas A\&M University}
  \city{College Station}
  \country{USA}
}

\renewcommand{\shortauthors}{Hangxiao Zhu et al.}

\begin{abstract}
  Communicating across cultures is inherently challenging, especially through culturally dense and ambiguous formats like memes. While people expect large language models (LLMs) hold promise for bridging such gaps, existing benchmark datasets often fail to capture the cultural context necessary for accurate interpretation. To address this, we introduce \textsc{MemeBridge}, a curated dataset centered on U.S.-originated memes, designed to capture two complementary perspectives: (1) how Chinese participants interpret these memes, and (2) how U.S. participants anticipate how people from other cultures might misunderstand them. Here, context refers to implicit cultural knowledge—background beliefs, norms, and shared assumptions that shape meme comprehension. The dataset was constructed via a multi-stage crowdsourcing pipeline with rigorous validation, including human agreement checks and GPT-based classification verification. Each meme is annotated with sentiment, emotion, cultural significance, and knowledge type, providing rich supervision for downstream tasks. Notably, we observe that the anticipated misunderstandings from U.S. participants are often inaccurate, highlighting the asymmetries in cultural understanding and the challenges of adopting perspectives beyond one's own. This bidirectional framing—focusing on both expression and perception—enables more nuanced benchmarking of cross-cultural comprehension. Our probing of multiple LLMs reveals that while models developed in different cultural contexts exhibit partial cross-cultural understanding, they often struggle with sophisticated interpretations. By contrast, fine-tuning with \textsc{MemeBridge} improves model performance, underscoring the value of culturally grounded resources for training and evaluating LLMs in globally diverse settings.
\end{abstract}

\ccsdesc[500]{Human-centered computing~Human computer interaction (HCI)}
\ccsdesc[500]{Applied computing~Sociology}

\keywords{Human Behavior Analysis, NLP Tools for Social Analysis}



\maketitle

\begin{spacing}{0.99}
\section{Introduction}
\input{introduction.tex}

\section{Related Work}
\input{related_work.tex}

\section{\textsc{MemeBridge} Dataset Construction}
\input{dataset.tex}

\section{Evaluating LLM's Cross-Cultural Meme Understanding}
\input{model_test.tex}

\section{Discussions}
\input{discussions.tex}

\section{Limitations}
\input{limitations.tex}
\end{spacing}


\bibliographystyle{ACM-Reference-Format}
\bibliography{citations}

\appendix
\label{sec:appendix}
\input{appendix.tex}

\end{document}

%% file: introduction.tex
\begin{figure}[t]
    \centering
    \includegraphics[width=0.46\textwidth]{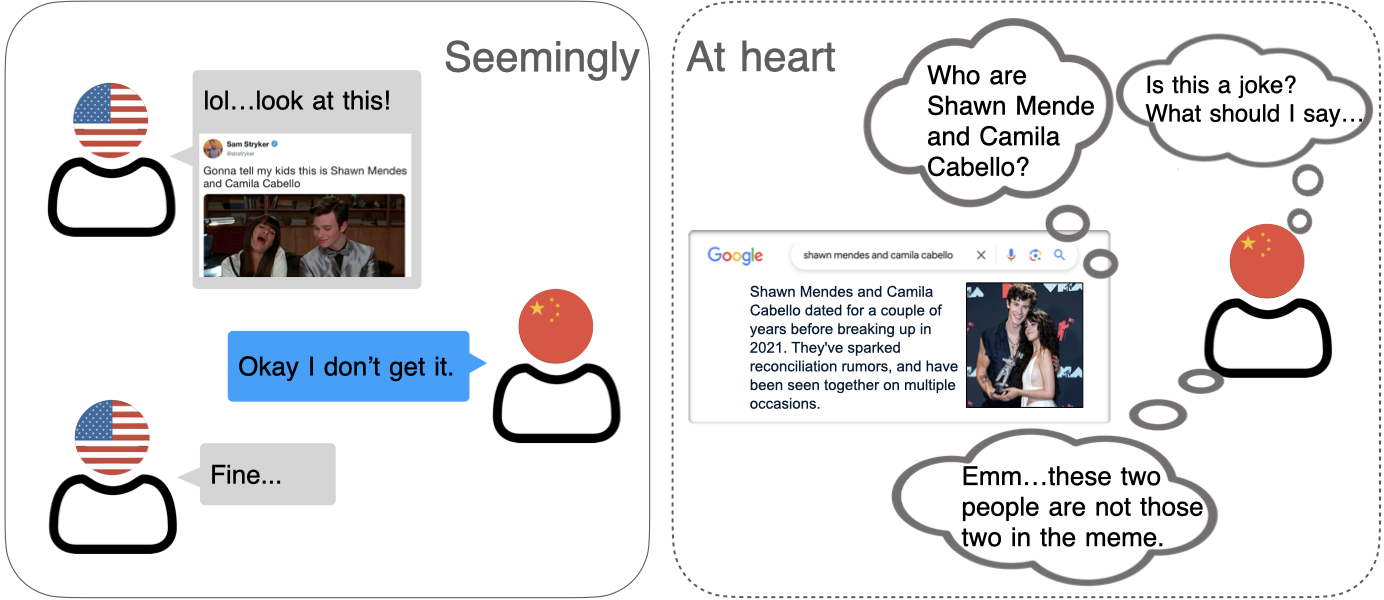}
    \Description{A meme image referenced in the caption.}
    \vspace{-0.5em}
    \caption{This meme humorously distorts history, exaggerating Mendes and Cabello’s relationship. A Chinese individual unfamiliar with the `Gonna Tell My Kids' meme format or U.S. pop culture may find it confusing, leading to awkward interactions.}
    \label{fig:intro}
\end{figure}

Memes serve as a form of speech act in digital communication, enabling Internet users to engage in social interactions through shared cultural references and semiotic cues~\cite{Grundlingh15032018}. However, memes are more than just visual humor; they function as cultural artifacts that reflect societal trends, linguistic variations, and generational differences. Their interpretation is often deeply rooted in a cultural context, making them susceptible to misinterpretation by individuals from different backgrounds~\cite{Mukhtar2024Memes}. For example, when Chinese individuals attempt to interpret memes originating in the United States, significant gaps can be seen with respect to humor, historical references, and societal norms. To investigate this issue, we conducted informal interviews with eight Chinese individuals currently residing in the United States. Many participants shared concerns about the potential for miscommunication when using memes, as illustrated by the following quote.
\begin{displayquote}
    \textit{"... Honestly, sometimes I worry about using memes incorrectly and accidentally causing awkwardness or conflicts with my (American) friends...."}
\end{displayquote}
Moreover, as global social media platforms continue to expand, this problem could become increasingly common, even among Chinese people not living in the United States. Following our interview, although Chinese individuals typically use platforms within their own cultural group, such as WeChat and Weibo, many of them also engage with global platforms like Twitter and Facebook, thereby being exposed to other cultures as well~\cite{Tsai02012017}. A lack of cultural familiarity can lead to unintended misunderstandings or misaligned social interactions, as in Figure~\ref{fig:intro}. Addressing these gaps is crucial for improving cross-cultural digital communication and mitigating the risk of misinterpretation in online discourse.

As large language models (LLMs) continue to advance, multimodal variants such as GPT-4o~\cite{hurst2024gpt}, Llama-3.2-Vision~\cite{dubey2024llama}, GLM-4V~\cite{glm2024chatglmfamilylargelanguage}, and Qwen-VL~\cite{bai2023qwenvlversatilevisionlanguagemodel} have become increasingly accessible to the general public. Given their ability to process and generate both textual and visual content, multimodal LLMs present a potential solution to bridge cultural gaps in communication, including the interpretation of memes. However, previous research on cultural knowledge bases~\cite{shi2024culturebank} has demonstrated that LLMs predominantly reflect Western-centric perspectives, making it challenging for non-Western audiences to fully understand culturally embedded content. Additionally, biases in training data and limitations in understanding the relationship between text and image can lead LLMs to generate skewed or inaccurate meme explanations~\cite{Zhong_2024_CVPR}. Despite the increasing sophistication of LLMs, these limitations raise concerns about whether they can effectively interpret and contextualize memes across cultures and help people understand memes from different countries, necessitating further investigation into their performance in cross-cultural meme understanding.

Several existing meme datasets have supported progress in meme understanding, captioning, and harmful content detection. However, these datasets are not designed to explicitly address cross-cultural interpretation. In contrast, our work focuses on bridging cultural gaps in meme understanding by introducing a bidirectional framing that captures both native-speaker interpretations and potential cross-cultural misunderstandings. In addition, we incorporate sentiment and emotion annotations grounded in cultural context, along with native-speaker verification to ensure interpretive fidelity. Table~\ref{tab:dataset_comparison} summarizes these key distinctions, highlighting how \textsc{MemeBridge} complements prior resources rather than serving as a direct substitute.

\begin{table}[t]
\centering
\small
\setlength{\tabcolsep}{3pt}
\renewcommand{\arraystretch}{1.05}
\resizebox{\columnwidth}{!}{%
\begin{tabular}{lcccc}
\toprule
\textbf{Feature} & \textbf{FigMemes} \cite{liu2022figmemes} & \textbf{MemeCap} \cite{hwang2023memecap} & \textbf{Multi3Hate} \cite{bui2024multi3hate} & \textbf{MemeBridge} \\
\midrule
Bidirectional Framing     & \xmark & \xmark & \xmark & \cmark \\
Sentiment Annotation      & \xmark & \xmark & \xmark & \cmark \\
Emotion Annotation        & \xmark & \xmark & \xmark & \cmark \\
Interpretation Caption    & \xmark & \cmark & \xmark & \cmark \\
Native-Speaker Annotation & \xmark & \cmark & \cmark & \cmark \\
\bottomrule
\end{tabular}%
}
\vspace{1em}
\caption{Comparison between \textsc{MemeBridge} and existing meme datasets.}
\vspace{-3em}
\label{tab:dataset_comparison}
\end{table}

In this paper, we investigate the ability of state-of-the-art LLMs to interpret U.S.-based memes, focusing on their capacity to provide explanations, detect sentiment, and identify emotions. To facilitate this study, we constructed \textsc{MemeBridge}, a carefully curated dataset consisting of memes contributed by native U.S. participants. Each meme entry includes explanations, potential misunderstandings that individuals from different cultural backgrounds might experience, and sentiment and emotion annotations. Using this dataset, we evaluate and fine-tune multiple LLMs to assess their effectiveness in cross-cultural meme interpretation.

Through extensive experiments, we have the following key observations: (1) The cultural gap in meme interpretation is bidirectional. Chinese individuals face challenges in understanding U.S. memes, with 58.8\% accuracy in determining their explanations, 45\% accuracy in labeling sentiment and 48.9\% accuracy in labeling emotions. Similarly, U.S. participants struggled to accurately predict how Chinese individuals misinterpret memes, as the misunderstandings proposed by U.S. participants often did not align with actual misconceptions held by Chinese participants. (2) Model performance is not strongly tied to the country of origin. Despite being developed by organizations based in either China or the United States, all evaluated LLMs exhibited comparable performance and behavioral patterns across tasks. This suggests that contemporary alignment and safety-tuning practices mitigate overt, origin-specific cultural biases, rather than embedding strong national perspectives. (3) LLMs demonstrate cultural awareness and adaptability. Explicitly instructing LLMs to adopt a specific cultural perspective significantly impacts their interpretative performance. All tested models exhibited performance differences when role-playing as U.S. participants or role-playing as Chinese participants, compared to the default setting. This suggests their ability to adjust to different cultural identities and perspectives.

%% file: related_work.tex

\textbf{Cultural Awareness in LLMs.}
The popularity and adoption of LLMs in various domains pose challenges and the need for cultural awareness~\cite{pawar2024survey,ramezani2023knowledge}. Existing studies have found that LLMs have biases in understanding cultural symbols, have different performances for different regional cultures, and are difficult to reach human levels~\cite{yao2024benchmarking}. For example, \citet{shi2024culturebank} have demonstrated that LLMs predominantly reflect Western-centric perspectives, making it challenging for non-Western audiences to fully understand culturally embedded content.
To address this, a growing number of studies~\cite{shi2024culturebank,zhao2024worldvaluesbench,li2024culture} have explored various aspects of integrating cultural understanding into LLMs, with the aim of bridging communication gaps and facilitating effective cross-cultural exchange. 
%
For example, \citet{nguyen2023extracting} has proposed Candle, an end-to-end approach to extracting cultural common sense knowledge from Web corpora on a large scale. 
Although these studies offer valuable information, many of them focused on machine translation with text information. More recently, the concept of image transcreation for cultural relevance acknowledges the need to adapt visual content for cultural appropriateness~\cite{khanuja2024image}, representing a crucial step towards bridging the gap between visual language understanding and cultural interpretation. 

\smallskip
\noindent \textbf{Cross-Cultural Understanding with Multimodal LLMs.}
Some recent works on multimodal LLMs highlight the challenges of adapting multimodal reasoning across diverse linguistic and cultural contexts. One major focus has been cultural adaptation in multimodal tasks, where researchers explore how models interpret visual and textual information differently across cultures, emphasizing the need for datasets that reflect such diversity~\cite{liu2021visually, li2023cultural}. Another key area is culturally influenced language inference, examining how cultural norms shape reasoning, particularly in tasks like natural language inference and figurative language understanding~\cite{huang2023culturally, kabra2023multi}. Additionally, work on humor, satire, and harmful content detection demonstrates the necessity of culturally aware AI, as humor and hate speech often rely on nuanced cultural context~\cite{nandy2024yesbut, bui2024multi3hate}. Collectively, these studies stress the importance of integrating cultural awareness into vision-language models to enhance their robustness and fairness in global applications.
These studies inspired us to investigate the cross-cultural understanding of memes, the dynamic and informal media circulating in online communities.

\smallskip
\noindent \textbf{Cross-Cultural Understanding and Evaluation of Memes.}
Several datasets have been collected to facilitate the understanding of memes. For example, \citet{zannettou2018origins} collected and analyzed 160 million images from four major online communities (i.e., Twitter, Reddit, 4chan's /pol/, and Gab), establishing a methodological framework for cross-platform meme tracking and analysis. 
FigMemes~\cite{liu2022figmemes} focuses on the identification of figurative language in political memes.
MCC~\cite{akhtarmemex} contains 3,400 memes and their contexts focusing on detecting explanatory evidence for memes.
MemeCap~\cite{hwang2023memecap} enables the evaluation of visual language models in the meme captioning task.
SemanticMemes~\cite{zhou2023social} highlights semantic clustering.
MemeMQA~\cite{agarwal2024mememqa} offers a multimodal question-answering framework for a better semantic explanation of memes.
Multi3hate~\cite{bui2024multi3hate} is designed for specific tasks such as context understanding and hate speech detection.
There studies enhanced the understanding and interpretation of memes but did not adequately address their understanding from a cross-cultural perspective. Our work collects data through crowdsourcing and requires participants to provide an explanation and possible misconceptions of each meme, as well as sentiment and emotion tags. 
Furthermore, we propose a novel cross-cultural evaluation design by prompting LLMs as people of different cultural backgrounds, enabling a more direct and quantitative assessment of cross-cultural misunderstandings. 


\begin{figure*}[htbp]
    \centering
    \includegraphics[width=\textwidth]{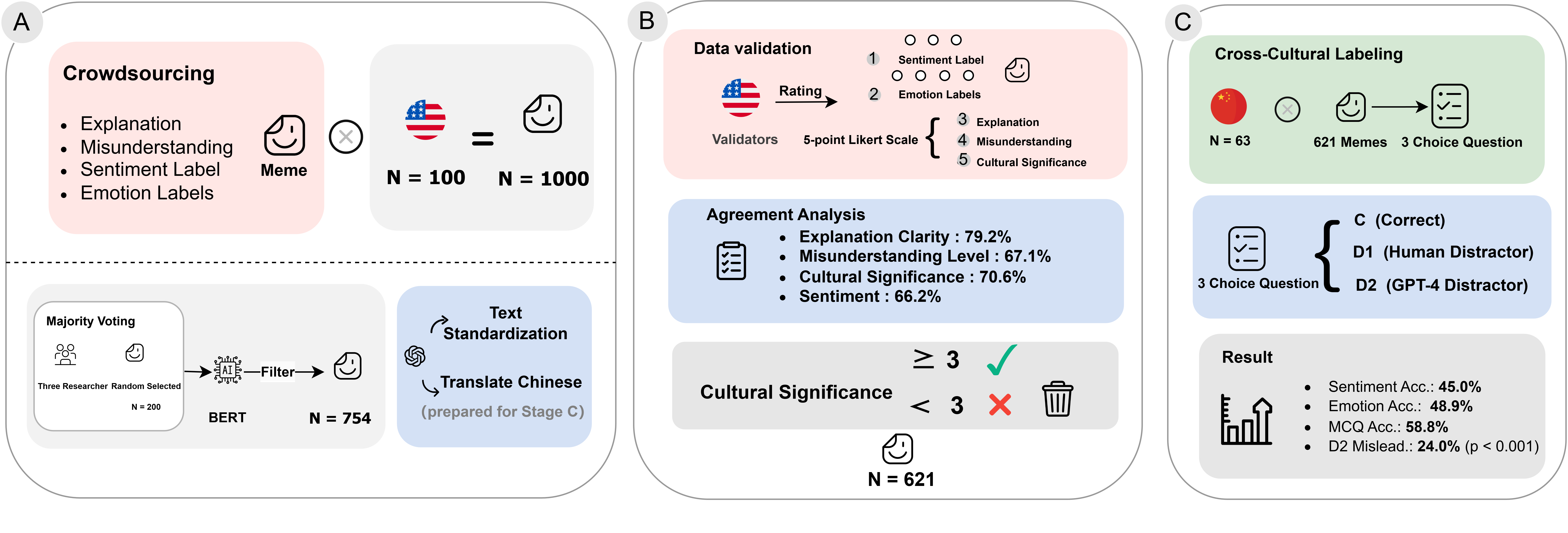}
    \Description{Pipeline.}
    \vspace{-2em}
    \caption{(A) Data Collection \& Cleaning: 200 memes randomly selected from 1000 collected and labeled by three researchers, followed by BERT-based refinement (N=754). (B) Data Validation: Participants labeled memes for sentiment and emotion. Explanation text, misunderstandings, and cultural significance were rated on a five-point Likert scale. Resulting in the final meme dataset (N=621) (C) Cross-Cultural Labeling: Chinese participants participated in meme interpretation tests.}
    \vspace{-0.5em}
    \label{fig:pipeline}
\end{figure*}

%% file: dataset.tex
To ensure high-quality data for cross-cultural meme understanding, we designed a three-stage crowdsourcing pipeline for dataset collection, validation, and cross-cultural testing. This structured process aims to systematically refine and validate the dataset while identifying cultural misunderstandings embedded in memes. Participant demographics are provided in Appendix~\ref{app:demographics}, while recruitment and compensation details are available in Appendix~\ref{app:recruitment}. The full research consent form is included in Appendix~\ref{app:consent}. The \textsc{MemeBridge} dataset is publicly available\footnote{\url{https://drive.google.com/drive/folders/152AN3iREfi71WThArmr8OcUWM5cV8YZy}}.

\subsection{Stage 1: Initial Data Collection}
\label{subsec:stage1}
The dataset construction process began with collecting a diverse set of memes and their interpretations from a U.S. crowd group consisting of 100 participants, recruited through Prolific. Each participant was asked to contribute 10 memes along with their personal \textit{explanations} and \textit{potential misunderstandings} they believed could arise for individuals from other cultural backgrounds, yielding a total of 1,000 data points. To ensure consistent interpretation, annotators were provided with detailed instructions, including an illustrative example: how non-U.S. participants might misinterpret the phrase ``Netflix and Chill'' as a literal movie invitation, not realizing its euphemistic meaning. This encouraged contributors to reflect on how culturally specific cues might be missed or misread. The full annotation prompt is included in Appendix~\ref{app:annotation-instructions}. In addition to these textual inputs, participants were asked to assign each submitted meme a sentiment label from $\{$positive, negative, neutral$\}$ and one or more emotion labels from $\{$sarcastic, humorous, offensive, motivational$\}$~\cite{sharma-etal-2020-semeval}. This approach ensures that the dataset captures not only the explicit meaning of memes but also the emotions and cultural context associated with them. Our crowdsourcing method aligns with prior efforts, such as~\citet{yin-etal-2022-geomlama}, which leveraged diverse participant contributions to collect geo-diverse commonsense knowledge.

To enhance data quality, we randomly selected 200 data points and had three researchers label them using a binary scheme, i.e. classifying each explanation and misunderstanding as either ``good'' or ``bad'' quality. Majority voting was used to determine the final label for each data point. This labeled dataset was then used to train a BERT-based classifier~\cite{devlin2019bertpretrainingdeepbidirectional} to filter out low-quality meme interpretations. To justify the classifier’s reliability and training sample sufficiency, our results in Appendix~\ref{appendix:bert-performance} show that 200 training samples are enough to achieve strong classification performance. Next, we conducted a linguistic complexity check, revealing that explanations contained an average of 26.12 words, while misunderstandings averaged 20.89 words. A Type-Token Ratio (TTR) analysis~\cite{richards1987type} further showed that explanations had an average TTR of 0.905, whereas misunderstandings had a slightly higher average of 0.928. To ensure high linguistic quality, we filtered out low-diversity data by computing the average TTR of each explanation and misunderstanding. Data points with an average TTR below 0.5 were discarded, retaining only those with sufficient lexical diversity. After applying these filtering steps, 754 data points met the quality criteria and were retained for further analysis.

Following these analyses, we used the GPT-4 API~\cite{openai2024gpt4technicalreport} to rewrite the original data, standardize the format, and improve grammatical accuracy while preserving semantic integrity. To ensure consistency, we applied a similarity scoring mechanism to compare the refined text with the original. Details of our prompt design and continuous monitoring of similarity scores to ensure successful rewriting are provided in the Appendix~\ref{rewrite}. Finally, we also leveraged GPT-4 to translate the dataset into Chinese, preparing it for comparative cross-cultural evaluation in Stage 3.

\subsection{Stage 2: Data Validation}
To ensure robustness and reliability, we conducted a validation process involving another group of 180 U.S. participants. However, some participants left early or failed the attention check. To maintain consistency, we ensured that each meme was reviewed and annotated by exactly four participants. This phase focused on measuring consistency and agreement among annotators to assess the quality of the collected explanations and misunderstandings. Participants were asked to assign sentiment and emotion labels to the memes to gauge consensus and alignment in interpretations. Moreover, the explanatory text, identified misunderstandings and cultural significance were evaluated using a five-point Likert scale, allowing us to quantify recognition and ensure the cultural relevance of our data. 

To assess agreement levels, we computed percent agreement for each meme across multiple dimensions: explanation clarity, misunderstanding level, cultural significance (all mapped to a three-level scale derived from the five-point Likert ratings), sentiment, and emotion. Agreement was determined by identifying the modal rating among the four annotators and calculating the proportion of annotators who assigned the same rating. Our results showed agreement rates of 79.2\% for explanation clarity, 67.1\% for misunderstanding level, 70.6\% for cultural significance, 66.2\% for sentiment, and between 75\% and 90\% for the four emotion labels. Based on these results, we filtered out memes with an aggregated cultural significance rating below 3 (on a five-point Likert scale, meaning that most annotators did not perceive them as culturally significant in the U.S. context). After filtering, 621 memes remained in the final dataset, and we assigned the aggregated sentiment and emotion labels to them for further analysis. 

We summarize the final distribution of sentiment and emotion labels in Table~\ref{tab:sentiment_emotions}. This shows a reasonably balanced sentiment distribution, with a diverse range of emotions well-represented.

\begin{table}[t]
\centering
\small
\setlength{\tabcolsep}{4pt}
\renewcommand{\arraystretch}{1.1}
\begin{tabular}{lrrrr}
\toprule
 & \textbf{Positive} & \textbf{Neutral} & \textbf{Negative} & \textbf{Total} \\
\midrule
\textbf{Sarcastic}     & 31  & 91  & 60  & 182 \\
\textbf{Humorous}      & 167 & 285 & 87  & 539 \\
\textbf{Motivational}  & 24  & 9   & 2   & 35  \\
\textbf{Offensive}     & 3   & 0   & 24  & 27 \\
\bottomrule
\end{tabular}
\vspace{1em}
\caption{Distribution of sentiment labels within each emotion category.}
\vspace{-3em}
\label{tab:sentiment_emotions}
\end{table}

\begin{figure}[t]
\centering
\includegraphics[width=0.8\linewidth]{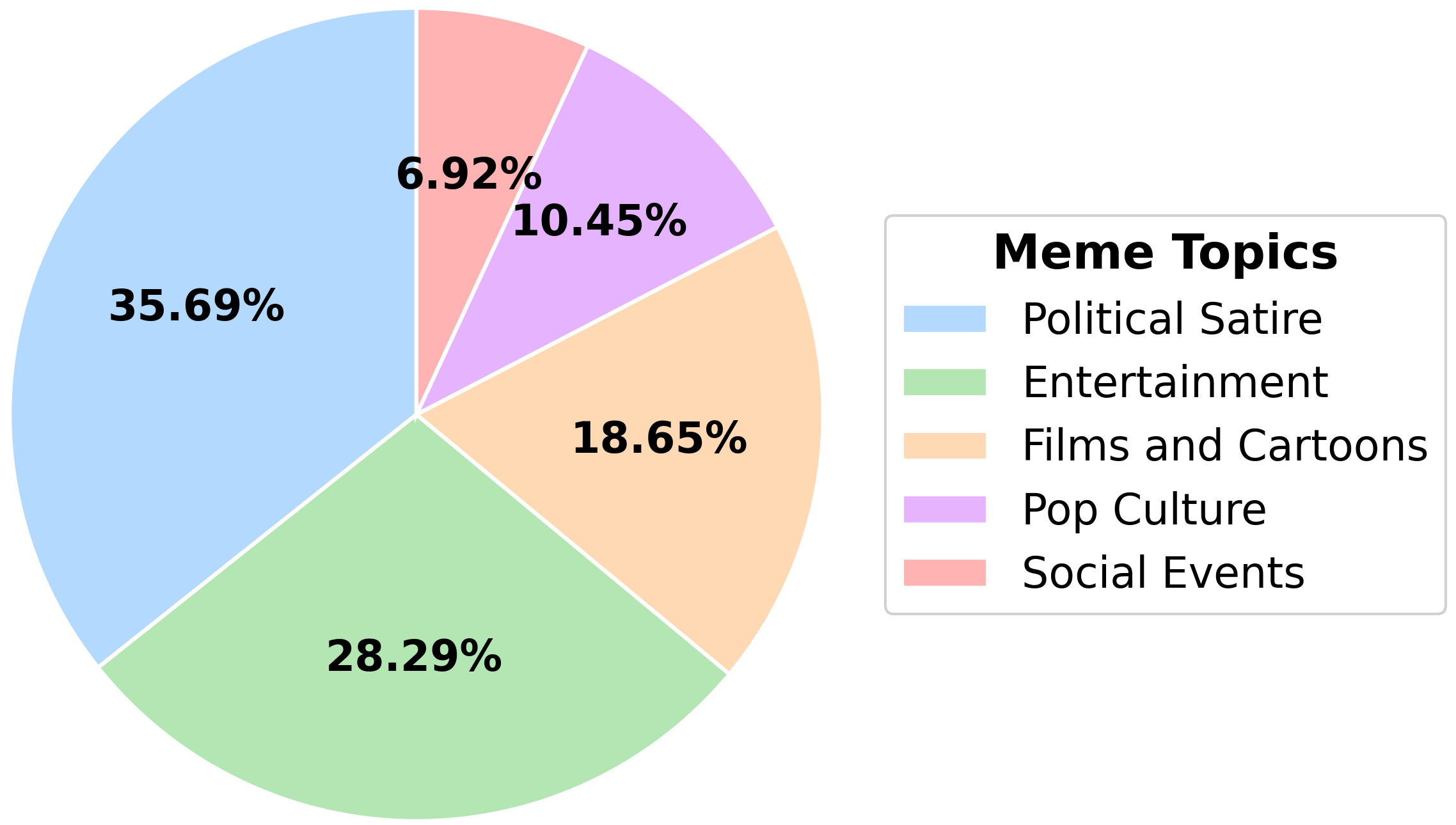}
\Description{A pie chart showing meme topics distribution.}
\vspace{-0.5em}
\caption{Distribution of meme topics in the dataset.}
\vspace{-1em}
\label{fig:pie-chart}
\end{figure}

To further analyze the dataset, we employed GPT-4 to classify each meme along two dimensions. The prompts used for these classification tasks are provided in Appendix~\ref{prompt:classification}. For topic labeling, we adopted five categories inspired by prior work such as MEMEX~\cite{akhtarmemex}: Political Satire, Entertainment, Films and Cartoons, Pop Culture, and Social Events. The distribution of our dataset across these categories is shown in Figure~\ref{fig:pie-chart}. Each meme was also categorized as requiring either cultural or general knowledge for correct interpretation. This binary classification was again performed using GPT-4o, with one label per meme. Table~\ref{tab:knowledge-type} shows the resulting distribution. Together, these results confirm that the dataset spans diverse topical domains and is rich in culture-specific content—consistent with our goal of facilitating research on cross-cultural meme understanding.

\begin{table}[t]
\centering
\small
\begin{tabular}{lc}
\toprule
\textbf{Knowledge Type} & \textbf{Number of Memes} \\
\midrule
Cultural Knowledge & 529 \\
General Knowledge  & 92  \\
\bottomrule
\end{tabular}
\vspace{1em}
\caption{Distribution of memes by required knowledge type.}
\vspace{-2em}
\label{tab:knowledge-type}
\end{table}

We conducted a separate human validation study to confirm the agreement level and assess the reliability of our automatic topic and knowledge-type annotations (general vs. cultural knowledge). We randomly sampled 200 memes from the dataset, and one author manually labeled each meme for both topic and knowledge type. These human-provided labels were then compared with GPT-4o’s automatic classifications. The agreement rates by category and Cohen’s Kappa are shown in Table~\ref{tab:topic_knowledge_agreement}. These results demonstrate strong alignment between human and GPT-4o annotations, with Cohen’s Kappa values of 0.89 for topic classification and 0.85 for knowledge-type classification. This high level of agreement provides strong empirical evidence for the reliability of automatic labeling in our dataset.

\begin{table}[t]
\centering
\small
\begin{tabular}{l c}
\multicolumn{2}{c}{\textbf{(Topic Classification)}} \\
\toprule
\textbf{Category} & \textbf{Agreement (\%)} \\
\midrule
Political Satire & 91.3 \\
Entertainment & 93.4 \\
Films and Cartoons & 93.3 \\
Pop Culture & 94.1 \\
Social Events & 83.3 \\
\midrule
\textbf{Cohen’s Kappa} & \textbf{0.89} \\
\bottomrule
\end{tabular}
\vspace{0.5em}
\begin{tabular}{l c}
\multicolumn{2}{c}{\textbf{(Knowledge Type Classification)}} \\
\toprule
\textbf{Label} & \textbf{Agreement (\%)} \\
\midrule
General vs. Cultural & 95.5 \\
\midrule
\textbf{Cohen’s Kappa} & \textbf{0.85} \\
\bottomrule
\end{tabular}
\vspace{1em}
\caption{Human-GPT-4o agreement on topic and knowledge-type classification (N=200).}
\vspace{-2em}
\label{tab:topic_knowledge_agreement}
\end{table}

\subsection{Stage 3: Cross-Cultural Assessment}
\label{subsec:stage3}
The final stage aimed to evaluate cross-cultural differences in meme interpretation by engaging 84 Chinese participants. After filtering out those who left early or failed the attention check, 63 participants remained, ensuring that each meme was reviewed by two individuals, resulting in 1,242 data points (621 memes$\times$2 reviews/meme). These participants provided sentiment and emotion labels, allowing us to compare their perceptions with those of the U.S. participants and identify potential cultural divergences. 

Additionally, participants were asked to complete multiple-choice questions constructed in the following way:
\textit{Explanations} were designated as the correct answer $C$, while \textit{potential misunderstandings} served as one distractor $D_1$. To introduce further variation, we employed GPT-4 to generate an additional distractor $D_2$ by providing the meme as input. The prompt used for generating $D_2$ is included in Appendix~\ref{prompt:MCQ}. This resulted in a three-choice question format $\{C, D_1, D_2\}$ for each meme.

\begin{figure}[t]
\centering
\includegraphics[width=\columnwidth]{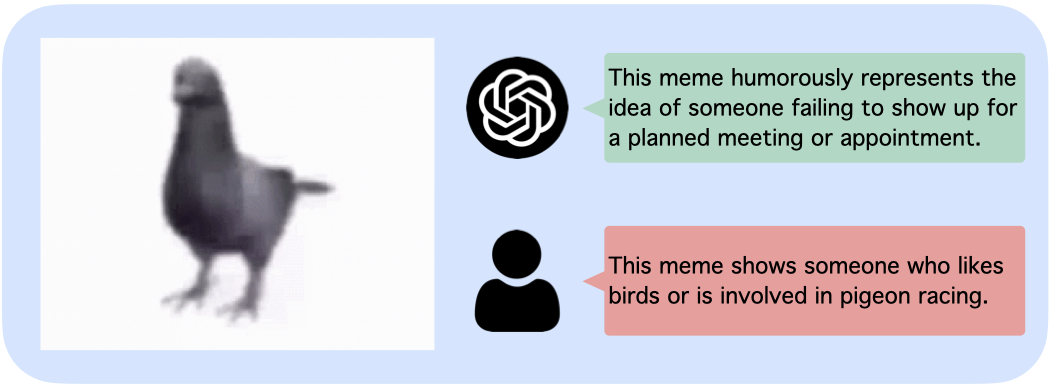}
\Description{A distractor example.}
\vspace{-1em}
\caption{Example comparison between GPT-4-generated and human-assumed \textbf{distractors} for two U.S. memes.}
\label{fig:distractor-example}
\vspace{-1em}
\end{figure}

Our assessment results demonstrated that Chinese participants struggled to accurately interpret sentiment in U.S.-centric memes, achieving only 45.0\% accuracy. Similarly, their ability to correctly identify emotions was limited, with an accuracy of 48.9\%. Most notably, their performance on the multiple-choice task was relatively low, with a correctness rate of just 58.8\%. As discussed in the introduction, these findings reinforce the existence of a cultural gap affecting meme comprehension. For the multiple-choice task, 17.1\% of all responses selected the human-assumed distractor ($D_1$), while 24.0\% selected the LLM-generated distractor ($D_2$), indicating that Chinese participants were significantly more misled by the LLM-generated distractor than the human-assumed distractor ($p < 0.001$). This supports observation 1: the cultural gap is bidirectional—just as Chinese participants struggle to interpret U.S. memes, U.S. participants may also have difficulty predicting how others will perceive their memes. To contextualize this difference, we include a comparison of LLM-generated and human-assumed distractors in Figure~\ref{fig:distractor-example}. Additional representative examples from the dataset are included in Appendix~\ref{app:dataset_example}.

To investigate whether misunderstandings by Chinese participants stem from a lack of cultural or general knowledge, we compared their performance across memes classified as requiring cultural versus general knowledge. Accuracy rates for three key tasks - multiple choice (MCQ), sentiment labeling, and emotion labeling - are shown in Table~\ref{tab:performance-comparison}. A significance marker (*) indicates $p < 0.05$. The sentiment classification task showed a significant performance gap, with participants performing worse on memes requiring cultural knowledge. Emotion labeling performance was also lower for cultural memes, with marginal significance. These results indicate that observed misunderstandings are more likely due to cultural barriers rather than a lack of general knowledge.

\begin{table}[t]
\centering
\small
\setlength{\tabcolsep}{4pt}
\renewcommand{\arraystretch}{1.1}
\begin{tabular}{lcc}
\toprule
\textbf{Task} & \textbf{Cultural Memes} & \textbf{General Memes} \\
\midrule
Multiple Choice (MCQ) & 59.69\% (0.491) & 54.04\% (0.500) \\
Sentiment Labeling    & 43.54\% (0.496) & 53.03\% (0.500)$^{*}$ \\
Emotion Labeling      & 47.76\% (0.500) & 55.56\% (0.498) \\
\bottomrule
\end{tabular}
\vspace{1em}
\caption{Chinese participant performance on memes requiring cultural vs.\ general knowledge. Values are mean accuracy with standard deviation in parentheses. $^{*}$ indicates statistically significant difference at $p < 0.05$.}
\vspace{-3em}
\label{tab:performance-comparison}
\end{table}

%% file: model_test.tex
With the dataset constructed, we aim to evaluate the performance of LLMs in meme interpretation, focusing specifically on four off-the-shelf multimodal LLMs: Qwen2.5-VL-3B~\cite{bai2023qwenvlversatilevisionlanguagemodel}, GLM-4V~\cite{glm2024chatglmfamilylargelanguage}, Llama-3.2-11B-Vision~\cite{dubey2024llama}, and GPT-4o~\cite{hurst2024gpt}\footnote{In the following discussion, we abbreviate these models to Qwen, GLM, LLaMA, and GPT for the ease of notation.}. Our goal is to assess the models' ability to generate human-like interpretations, accurately detect the sentiment and emotions conveyed by memes, and evaluate their adaptability to different cultural perspectives. All prompts used for model evaluation are provided in Appendix~\ref{prompt:model-test}.

\subsection{Assessment on LLMs}
First, we evaluated LLMs under the same test conditions as Chinese participants in Section~\ref{subsec:stage3}. Our findings indicate that while GPT consistently outperforms Chinese participants in interpreting U.S. memes, the other models exhibit specific weaknesses, as shown in Table~\ref{tab:model_vs_chnppl}. Notably, for sentiment classification, Qwen, GLM, and LLaMA all performed worse than Chinese participants, indicating that recognizing sentiment is inherently subtle and remains an open problem~\cite{10596638}.

In contrast, for emotion detection, Qwen and GLM significantly outperformed Chinese participants, indicating the potential of these models to assist non-native speakers in understanding emotions conveyed in U.S. memes. However, LLaMA performed consistently worse than both the other models and human participants. This could be attributed to the differences in dataset curation—while Qwen, GLM, and GPT are developed by companies with proprietary, curated training data, LLaMA is trained predominantly on open-source datasets, which may lack diversity, fine-grained annotations, or up-to-date meme-related content. Consequently, its performance on specialized tasks such as sentiment and emotion detection is notably lower.

\begin{figure}[t]
    \centering
    \includegraphics[width=\linewidth]{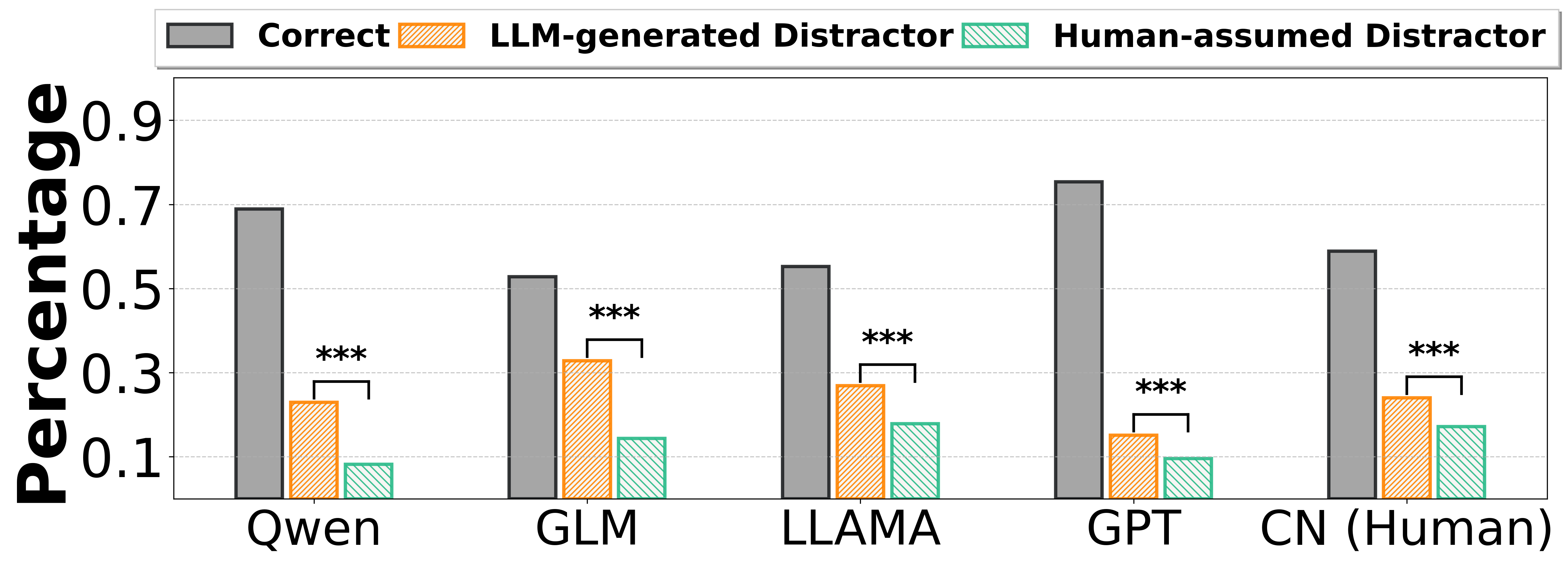}
    \Description{Model Choice Distribution.}
    \vspace{-1.5em}
    \caption{Comparing the distribution of answer choices across different LLMs and Chinese participants (CN (Human)).Significance indicators (*: p < 0.05, **: p < 0.01, ***: p < 0.001) above pairs of distractor bars show whether participants significantly favored one type of distractor over the other when making incorrect choices. }
    \vspace{-1em}
    \label{fig:choice}
\end{figure}

Further analysis of multiple-choice answers by different models revealed an interesting pattern: LLMs, similar to Chinese participants, tend to prefer LLM-generated distractors over human-assumed distractors, as shown in Figure~\ref{fig:choice}. This observation suggests that while LLMs can effectively understand U.S. memes, their errors align with `real' misunderstandings experienced by Chinese participants. This finding underscores the potential of LLMs in modeling cultural misinterpretations and highlights the dual nature of the cultural gap—both in interpreting and anticipating meme misunderstandings across cultures.

Further analysis of the multiple-choice responses revealed a consistent association: both LLM judges and our Chinese participants more frequently selected LLM-generated distractors than human-assumed distractors, as shown in Figure~\ref{fig:choice}. Importantly, this result is descriptive—we do not claim that LLM-generated distractors are inherently better or that their higher selection rate reflects a causal effect. In light of findings from the LLM-as-a-Judge literature on potential self-enhancement or source bias and other confounds (e.g., option position, verbosity/length, or implicit model cues)~\cite{gu2025surveyllmasajudge}, we interpret the pattern conservatively. Nonetheless, the alignment suggests that when LLMs make errors, they may do so in ways that resemble misunderstandings observed among Chinese participants, supporting the view that cultural gaps are bidirectional—not only in interpreting memes, but also in anticipating how they may be misread across cultures.

\begin{table}[t]
\small
\centering
\begin{tabular}{lccccc}
\toprule
& \textbf{Qwen} & \textbf{GLM} & \textbf{LLaMA} & \textbf{GPT} & \textbf{CN (Human)} \\
\midrule
MCQ & \textcolor[rgb]{0,0.502,0}{\uline{68.8\%}} & \textcolor{red}{\uline{52.8\%}} & \textcolor{red}{55.2\%} & \textcolor[rgb]{0,0.502,0}{\uline{75.4\%}} & 58.8\% \\
Sent & \textcolor{red}{40.4\%} & \textcolor{red}{\uline{37.7\%}} & \textcolor{red}{\uline{35.3\%}} & \textcolor[rgb]{0,0.502,0}{\uline{54.2\%}} & 45.0\% \\
Emo & \textcolor[rgb]{0,0.502,0}{\uline{65.1\%}} & \textcolor[rgb]{0,0.502,0}{\uline{64.2\%}} & \textcolor{red}{\uline{32.4\%}} & \textcolor[rgb]{0,0.502,0}{\uline{84.3\%}} & 48.9\% \\
\bottomrule
\end{tabular}
\vspace{1em}
\caption{Comparing the performance of different LLMs with Chinese participants on Meme Understanding. This table presents the accuracy of 4 different LLMs and Chinese annotators across three tasks: Multiple choice questions selection (MCQ), Sentiment labeling (Sent), and Emotion labeling (Emo). Underlined values indicate statistically significant differences from Chinese annotators (p < 0.05).}
\vspace{-3em}
\label{tab:model_vs_chnppl}
\end{table}

\subsection{Detection of LLMs' Potential Bias}
\begin{table*}[t]
\centering
\resizebox{\textwidth}{!}{
\begin{tabular}{l|ccc|ccc|ccc|ccc}
\hline
& \multicolumn{3}{c|}{\textbf{Qwen}}                & \multicolumn{3}{c|}{\textbf{GLM}}                 & \multicolumn{3}{c|}{\textbf{LLaMA}}               & \multicolumn{3}{c}{\textbf{GPT}}                  \\ 
& \default{DEF} & \usa{US-RP}                                 & \chn{CN-RP}                              & \default{DEF} & \usa{US-RP}                                 & \chn{CN-RP}                              & \default{DEF} & \usa{US-RP}                                 & \chn{CN-RP}                              & \default{DEF} & \usa{US-RP}                                 & \chn{CN-RP}                              \\ 
\hline
$\text{Sim}_\text{original}$  & 0.484   & \textcolor[rgb]{0,0.502,0}{0.492↑}  & \textcolor{red}{0.468↓}             & 0.429   & \textcolor[rgb]{0,0.502,0}{0.449↑}  & \textcolor{red}{0.428↓}             & 0.455   & \textcolor[rgb]{0,0.502,0}{0.467↑}  & \textcolor{red}{0.442↓}             & 0.460   & \textcolor[rgb]{0,0.502,0}{0.505↑}  & \textcolor[rgb]{0,0.502,0}{0.471↑}  \\ 
$\text{Sim}_\text{formatted}$ & 0.582   & \textcolor[rgb]{0,0.502,0}{0.600↑}  & \textcolor{red}{0.554↓}             & 0.479   & \textcolor[rgb]{0,0.502,0}{0.536↑}  & \textcolor{red}{0.475↓}             & 0.546   & \textcolor[rgb]{0,0.502,0}{0.566↑}  & \textcolor{red}{0.536↓}             & 0.499   & \textcolor[rgb]{0,0.502,0}{0.605↑}  & \textcolor[rgb]{0,0.502,0}{0.554}  \\ 
\hline
$\text{Acc}_\text{MCQ}$               & 68.8\%  & \textcolor{red}{67.9\%↓}            & \textcolor[rgb]{0,0.502,0}{69.5\%↑} & 52.8\%  & \textcolor{red}{46.7\%↓}            & \textcolor[rgb]{0,0.502,0}{52.9\%↑} & 55.2\%  & \textcolor{red}{47.8\%↓}            & \textcolor{red}{44.8\%↓}            & 75.4\%  & \textcolor{red}{72.7\%↓}            & \textcolor{red}{72.3\%↓}            \\ 
$\text{Acc}_\text{Sent}$              & 40.4\%  & \textcolor[rgb]{0,0.502,0}{46.2\%↑} & \textcolor[rgb]{0,0.502,0}{47.0\%↑} & 37.7\%  & \textcolor{red}{33.2\%↓}                             & \textcolor[rgb]{0,0.502,0}{38.3\%↑} & 35.3\%  & \textcolor[rgb]{0,0.502,0}{40.1\%↑} & \textcolor[rgb]{0,0.502,0}{39.0\%↑} & 54.2\%  & \textcolor[rgb]{0,0.502,0}{55.0\%↑} & \textcolor{red}{51.9\%↓}            \\ 
$\text{Acc}_\text{Emo}$               & 65.1\%  & \textcolor[rgb]{0,0.502,0}{79.4\%↑} & \textcolor[rgb]{0,0.502,0}{74.3\%↑} & 64.2\%  & \textcolor[rgb]{0,0.502,0}{67.5\%↑} & \textcolor[rgb]{0,0.502,0}{77.2\%↑} & 32.4\%  & \textcolor[rgb]{0,0.502,0}{42.2\%↑} & \textcolor[rgb]{0,0.502,0}{41.7\%↑} & 84.3\%  & \textcolor[rgb]{0,0.502,0}{84.5\%↑} & \textcolor{red}{82.6\%↓}            \\ 
\hline
\textbf{PS}         & 2.887   & \textcolor[rgb]{0,0.502,0}{3.192↑(0.305)}                              & \textcolor[rgb]{0,0.502,0}{3.049↑(0.162)}                              & 2.595   & \textcolor[rgb]{0,0.502,0}{2.663↑(0.068)}                              & \textcolor[rgb]{0,0.502,0}{2.821↑(0.226)}                              & 2.135   & \textcolor[rgb]{0,0.502,0}{2.321↑(0.186)}                              & \textcolor[rgb]{0,0.502,0}{2.232↑(0.097)}                              & 3.242   & \textcolor[rgb]{0,0.502,0}{3.385↑(0.143)}                              & \textcolor[rgb]{0,0.502,0}{3.245↑(0.003)}     \\
\hline
\end{tabular}
}
\vspace{1em}
\caption{Performance of LLMs across cultural settings and evaluation metrics. This table shows the performance of different LLMs under corresponding prompting strategies: \default{DEF} (Default Setting), \usa{US-RP} (US RolePlaying), and \chn{CN-RP} (Chinese RolePlaying). Performance is evaluated using metrics including similarity with original data (Sim$_{\text{original}}$) and formatted (Sim$_{\text{formatted}}$) text, Accuracy on Multiple choice questions selection (Acc$_{\text{MCQ}}$), Sentiment labeling (Acc$_{\text{Sent}}$), and Emotion labeling (Acc$_{\text{Emo}}$) tasks, along with a \textit{Performance Score} (PS). For statistical significance test and pairwise comparisons across cultural settings, please refer the details in Table 3.}
\label{tab:role_play}
\vspace{-2em}
\end{table*}
To evaluate cross-cultural adaptation and cultural awareness, we designed experiments to identify potential biases in model training that may result in cultural tendencies. We selected four models for comparison: Qwen2.5-VL and GLM-4V, Chinese developed open-source models; Llama-3.2-Vision, a U.S.-based open-source model; and GPT-4o, a widely regarded state-of-the-art closed-source model. Each model was tested under three conditions: the default setting \default{DEF}, an explicit prompt instructing the model to respond as a native US person \usa{US-RP}, and another instructing it to respond as a native Chinese \chn{CN-RP} \footnote{RP denotes \emph{role-playing}.}. Under each condition, the model first completed the same test as in Section~\ref{subsec:stage3}, and we measured their performance on those tasks. Then, in a fresh session, the models were instructed to generate an explanation for this meme. We compared these explanations with both the original (crowdworker-provided) explanations and the formatted (LLM-rewritten) explanations (both obtained in Section~\ref{subsec:stage1}), using cosine similarity scores to quantify textual alignment. The test results are presented in Table~\ref{tab:role_play}. Across all models, performance was highest under the \usa{US-RP} condition. Additionally, LLM-generated explanations exhibited higher similarity to the formatted explanations than to the original ones. This aligns with expectations, as LLM outputs tend to be more structured and formal, whereas crowdworker-written explanations exhibit more variability in grammar and vocabulary.

To summarize overall model performance across five evaluation metrics, we introduce a simple aggregate measure called the \textit{Performance Score} (PS), shown in Table~\ref{tab:role_play}. The score was computed by grouping the two similarity comparisons into a single task, along with three classification tasks: multiple choice questions selection (MCQ), sentiment labeling (Sent), emotions labeling (Emo):
\begin{align}
    \text{PS} &= (\text{Sim}_\text{original} + \text{Sim}_\text{formatted}) \nonumber \\
    &\quad + \text{PS}_\text{MCQ} + \text{PS}_\text{Sent} + \text{PS}_\text{Emo}.
\end{align}
Here, $\text{Sim}_\text{original}$ and $\text{Sim}_\text{formatted}$ represent the cosine similarity between the model-generated explanation and the original crowdworker explanation, and the GPT-4 generated explanation, respectively. $\text{Sim}_\text{formatted}$ serves as a complementary, style-controlled similarity measure. Both similarity terms are computed in parallel and equally weighted for all models, ensuring that the aggregate score remains model-fair while capturing semantic alignment under both original and controlled textual forms. $\text{PS}_\text{MCQ}$, $\text{PS}_\text{Sent}$, and $\text{PS}_\text{Emo}$ correspond to the normalized performance scores for the multiple-choice, sentiment, and emotion classification tasks. Each component is scaled to ensure comparable weight, allowing the PS to serve as a compact but interpretable summary of overall model behavior across both generative and classification tasks. Full details on score normalization and weighting are provided in Appendix~\ref{app:ps}. 

Note that for emotion labeling, a prediction is marked correct if the ground-truth labels are a subset of the predicted labels. We acknowledge that the subset-based evaluation rule for emotion labeling could raise concerns about degenerate solutions—such as models predicting all available emotion labels to maximize correctness. However, this risk was mitigated by design: the evaluation criteria were not disclosed to either models or human participants, making strategic label inflation unlikely. Furthermore, our empirical analysis (see Appendix~\ref{appendix:emotion-label-distribution}) confirmed that such behavior did not occur. Most predictions included only 1–2 emotion labels, and no response—model-generated or human—selected all four. The expected number of labels per response ranged from 1.27 to 1.63, far below the theoretical maximum of 4. These results validate the realism of model behavior under the subset-based metric and support its robustness for evaluating multi-label emotion classification.

While we report all sub-metrics individually in Table~\ref{tab:role_play}, we found that examining them in isolation can make it tedious to interpret trends across models and prompt settings. For example, the GLM model shows improved similarity scores under the \usa{US-RP} condition but performs worse on multiple-choice and sentiment classification compared to the default setting. Under \chn{CN-RP}, the pattern reverses: accuracy scores improve, while similarity scores decline. This contradictory behavior makes it difficult to determine which persona setting is most effective for the model overall. The PS provides a compact, interpretable summary to address this challenge. It enabled us to observe consistent trends—such as models performing best under one of the role-play conditions—without obscuring the underlying metric-level variation. Importantly, the PS is not intended to replace individual metrics but to serve as a complementary tool for high-level comparison.

Across all models, the performance score improved when models were explicitly instructed to adopt either the \usa{US-RP} or \chn{CN-RP} perspective, compared to the \default{DEF} condition. This finding suggests that role-playing prompts significantly impact LLMs' interpretative accuracy and their alignment with cultural contexts.

Pairwise significance tests (see Table~\ref{tab:culture_diff}) reveal that \usa{US-RP} consistently improves on \default{DEF} in a statistically significant manner in key metrics, and similar improvements are observed when comparing \chn{CN-RP} to \default{DEF} for most metrics. Notably, when directly comparing \usa{US-RP} and \chn{CN-RP}, the \usa{US-RP} condition generally outperforms. Overall, based on PS, the performance ranking for Qwen, LLaMA, and GPT follows the order: \usa{US-RP} > \chn{CN-RP} > \default{DEF}. These results indicate that persona-conditioned prompting can influence model behavior in ways that reflect alignment with cultural context. When the prompted identity matches the cultural background of the meme content—as in the \usa{US-RP} setting—models tend to show improved interpretability. In contrast, prompts that are less aligned with the content's origin—such as \chn{CN-RP}—are associated with relatively lower performance, though often still above the default setting. This suggests that models exhibit context-sensitive behavior when guided by role-specific instructions.

\begin{table*}
\centering
\resizebox{0.75\textwidth}{!}{
\begin{tabular}{c|c|c|c}
\hline
& \default{DEF} vs. \usa{US-RP} & \default{DEF} vs. \chn{CN-RP} & \usa{US-RP} vs. \chn{CN-RP} \\
\hline
\raisebox{-.2\height}{\textbf{Qwen}} 
    & \usa{$\text{Acc}_\text{Emo}$} & \default{$\text{Sim}_\text{formatted}$} \chn{$\text{Acc}_\text{Emo}$} \chn{$\text{Acc}_\text{Sent}$} & \usa{$\text{Sim}_\text{original}$} \usa{$\text{Sim}_\text{formatted}$} \\
\raisebox{-.2\height}{\textbf{GLM}} 
    & \usa{$\text{Sim}_\text{original}$} \usa{$\text{Sim}_\text{formatted}$} & \chn{$\text{Acc}_\text{Emo}$} & \usa{$\text{Sim}_\text{original}$} \chn{$\text{Acc}_\text{Emo}$} \\
\raisebox{-.2\height}{\textbf{LLaMA}} 
    & \usa{$\text{Sim}_\text{formatted}$} \usa{$\text{Acc}_\text{Emo}$} & \chn{$\text{Acc}_\text{Emo}$} & \usa{$\text{Sim}_\text{original}$} \usa{$\text{Sim}_\text{formatted}$} \\
\raisebox{-.2\height}{\textbf{GPT}} 
    & \usa{$\text{Sim}_\text{original}$} \usa{$\text{Sim}_\text{formatted}$} & \chn{$\text{Sim}_\text{formatted}$} & \usa{$\text{Sim}_\text{original}$} \usa{$\text{Sim}_\text{formatted}$} \\
\hline
\end{tabular}
}
\vspace{1em}
\caption{Metrics with significant differences across cultural settings for various LLMs (p < 0.05). For example, for the Qwen2.5-VL model, \usa{$\text{Acc}_\text{Emo}$} in the \default{DEF} vs. \usa{US-RP} column indicates significantly better performance of \usa{US-RP} on the emotion labeling task compared to \default{DEF}.}
\vspace{-2em}
\label{tab:culture_diff}
\end{table*}

\subsection{Fine-tuning}
To further validate the effectiveness of our dataset, we conducted fine-tuning experiments. The dataset was split into 70\% for fine-tuning, 15\% for validation, and 15\% for testing. We fine-tuned three models via their respective APIs: Qwen-2.5-VL-3B (developed by Alibaba), GLM-4V (by Zhipu AI), and GPT-4o (by OpenAI). Qwen-2.5-VL-3B has a publicly known parameter size of 3B. While the exact sizes of GLM-4V and GPT-4o are not disclosed, it is reasonable to assume that both are substantially larger than 9B parameters, especially in the case of GLM-4V, where the Zhipu AI API likely provides access to a more powerful variant than the open-sourced 9B model.

These models were evaluated on the same set of tasks: semantic similarity checking, multiple-choice question answering, sentiment labeling, and emotion classification. Overall, fine-tuning led to performance improvements across most tasks. However, an exception was observed with GPT in emotion classification, where accuracy dropped significantly from 87.1\% to 61.1\% on the test set. This decline may be attributed to overfitting, as the base GPT model already demonstrated strong performance prior to fine-tuning. Meanwhile, performance improvements were still observed in other tasks where fine-tuned GPT had not originally excelled. A similar trend was noted for Qwen, where its performance in generating explanations and multiple-choice question answering declined slightly after fine-tuning. Notably, this model initially outperformed the others in these tasks. However, fine-tuning resulted in substantial improvements in sentiment and emotion classification—areas where the base Qwen model had previously struggled.

These results suggest that while our dataset is effective in enhancing LLMs' capabilities in particularly intricate and niche tasks, the extent of improvement may depend on the pre-existing strengths of each model. Models that initially performed poorly, such as those struggling with sentiment classification, exhibited more noticeable improvements after fine-tuning, suggesting that our dataset is particularly beneficial for models with weaker prior knowledge and could possibly enhance their ability to interpret culturally relevant content. Conversely, models that were already strong in specific tasks, such as GPT in emotion classification, may experience diminishing returns or even degradation in performance due to overfitting. The graphs showing performance change are in Appendix~\ref{app:ft}.

%% file: discussions.tex
\subsection{The Bidirectional Cultural Gap and Usage of LLMs}
Our findings indicate that Chinese participants exhibited relatively low accuracy in multiple-choice question answering, sentiment labeling, and emotion classification. As shown in Table~\ref{tab:model_vs_chnppl}, they were frequently outperformed by LLMs, confirming one direction of the cultural gap: Chinese participants face challenges in understanding U.S. memes.

Additionally, when Chinese participants made errors in the multiple-choice task, they were more likely to select LLM-generated distractors rather than the human-assumed misunderstandings. This pattern was also observed in LLM testing. This suggests that U.S. participants’ assumptions don’t always reflect how Chinese individuals interpret memes. While LLMs can, to some extent, attempt to fathom out Chinese people's thought processes. This confirms the other direction of the cultural gap: U.S. participants may struggle to accurately anticipate how Chinese individuals interpret their memes.

These findings highlight an important application of LLMs beyond assisting Chinese participants in understanding U.S. memes. LLMs can also be leveraged to help U.S. participants anticipate potential misinterpretations of their shared content, allowing them to better understand how their messages might be perceived by individuals from different cultural backgrounds. In practical applications, this could help reduce misunderstandings, mitigate awkwardness, and prevent unintended conflicts in intercultural exchanges.

\subsection{Roleplaying Effects on LLMs}
Based on our experiment results, explicitly instructing LLMs to engage in role-playing can significantly enhance their performance on certain metrics, revealing that LLMs are aware of different cultural settings. This suggests that LLMs can adjust their behavior when guided to adopt a particular cultural perspective. However, the degree of improvement varies across different tasks and models, indicating that LLMs' underlying understanding may still be limited by their training data and pre-existing biases.

Interestingly, LLMs interpret U.S. memes better when prompted to act as native Chinese speakers compared to their default setting—but not as well as when role-playing as native U.S. speakers. This suggests that alignment has reduced cultural bias, likely to avoid favoring specific groups. While their stronger performance as English speakers reflects their English-heavy training data, alignment appears to suppress these biases. In some cases, this even lowers performance when mimicking non-U.S. perspectives. Overall, LLMs show not just reduced bias, but an ability to adapt culturally when explicitly instructed.

%% file: limitations.tex
While our study provides valuable insights into the role of LLMs in cross-cultural meme interpretation, several limitations should be acknowledged. The dataset size remains relatively small due to the high cost and logistical complexity of crowdsourcing raw meme data along with rich, structured annotations. As detailed in Appendix~\ref{app:recruitment}, the entire data collection process spanned approximately 40 days and incurred a cost of around \$2000. Despite these constraints, our multi-stage data collection pipeline has proven reliable and scalable for future expansion. The relatively small dataset size may negatively impact LLM fine-tuning, potentially leading to overfitting. While fine-tuning has generally improved model performance, certain tasks, such as emotion classification in GPT-4o, exhibited performance degradation, likely due to overfitting to the limited data.

Besides, although we identified a bidirectional cultural gap, our study did not validate its reversal—where Chinese participants provide memes and U.S. participants attempt to interpret them. It remains an open question whether Chinese participants would also struggle to predict potential misunderstandings by U.S. participants and how challenging U.S. participants would find it to interpret Chinese memes. Investigating this aspect would provide a more comprehensive understanding of cross-cultural meme interpretation.

Third, our study focuses exclusively on Chinese and U.S. cultural contexts, leaving out other linguistic and cultural backgrounds that may exhibit distinct patterns in meme interpretation. Future work should extend this research to a broader range of cultural settings to explore whether similar bidirectional gaps exist across other regions and communities.

Finally, we clarify that the ``context'' mentioned in our work refers to implicit cultural knowledge—the background beliefs, shared assumptions, and interpretive frameworks that shape how individuals understand memes. Although not explicitly embedded in text or image, this form of context plays a critical role in meme comprehension. Related to this, we also acknowledge that the potential misunderstandings collected in Stage 1 have limited predictive accuracy. These misunderstandings—imagined by U.S. participants—frequently failed to match actual responses from Chinese participants. We view this not as a flaw, but as a reflection of a deeper cultural asymmetry: that individuals often struggle to simulate interpretations from another cultural background. This limitation is amplified by the inherently subjective nature of misunderstanding, especially in ambiguous, culturally nuanced formats like memes. As misunderstandings are not objectively verifiable ground truths, it is difficult to collect reliable imagined misunderstandings through one-shot prompts. We hope future work can explore improved or interactive methods for eliciting these cross-cultural mismatches more accurately.

%% file: appendix.tex
\section{Participant Demographics}
\label{app:demographics}
To better understand the background of our crowdsourcing participants, we summarize below the key demographic distributions for both Stage 1 (initial data collection) and Stage 2 (validation). The Table~\ref{tab:demographics_summary} highlights sex, ethnicity, and employment status.

\newcolumntype{C}[1]{>{\centering\arraybackslash}p{#1}}

\begin{table}[t]
\centering
\scriptsize
\setlength{\tabcolsep}{2.5pt}
\renewcommand{\arraystretch}{0.95}
\resizebox{\linewidth}{!}{
\begin{tabular}{cC{1.5cm}p{3.0cm}p{3.2cm}}
\toprule
\textbf{Stage} & \textbf{Sex (F/M)} & \textbf{Ethnicity} & \textbf{Employment} \\
\midrule
Stage 1 & 38 / 62 &
Black (45), White (40), Mixed (12), Asian (2), Other (1) &
Full-time (50), Part-time (26), Unemployed (7), Other (15), Not in paid work (2) \\
\midrule
Stage 2 & 107 / 73 &
White (119), Black (35), Mixed (10), Asian (10), Other (6) &
Full-time (71), Part-time (26), Unemployed (19), Not in paid work (12), Other (50), Starting soon (2) \\
\bottomrule
\end{tabular}
}
\vspace{1em}
\caption{Participant demographics in Stage 1 and Stage 2.}
\vspace{-2em}
\label{tab:demographics_summary}
\end{table}

\section{Participant Recruitment and Compensation}
\label{app:recruitment}
We used different recruitment strategies across stages. \textbf{Stage 1} (data collection) and \textbf{Stage 2} (validation) participants were recruited via Prolific with U.S.-based screening and demographic diversity controls. \textbf{Stage 3} (cross-cultural testing) participants were recruited through digital flyers shared via university mailing lists and social media groups targeting Chinese international communities. All participants were paid at \$10/hour. Data collection lasted $\sim$40 days, with total recruitment/compensation costs of $\sim$\$2000.

\section{Research Study Consent Form}
\label{app:consent}
\textbf{Purpose:} Study cross-cultural understanding of internet memes.

\noindent\textbf{Procedure:} Tasks vary by stage and may include demographic questions and meme interpretation activities. Stage 1: upload $\ge$10 U.S. memes and provide cultural context, potential misunderstandings, sentiment, and emotion labels. Stage 2: validate meme interpretations and rate explanation/misunderstanding quality. Stage 3: answer multiple-choice interpretation questions for U.S. memes and label sentiment/emotions.

\noindent\textbf{Data Use \& Confidentiality:} Responses are anonymous (no personally identifiable information collected) and stored securely for research use, including potential future studies.

\noindent\textbf{Risks/Benefits \& Voluntary Participation:} No anticipated risks beyond everyday online content exposure. Participation is voluntary; you may skip questions or withdraw at any time without penalty.

\section{Annotation Instructions}
\label{app:annotation-instructions}
\begin{figure}[t]
  \centering
  \includegraphics[width=0.6\linewidth]{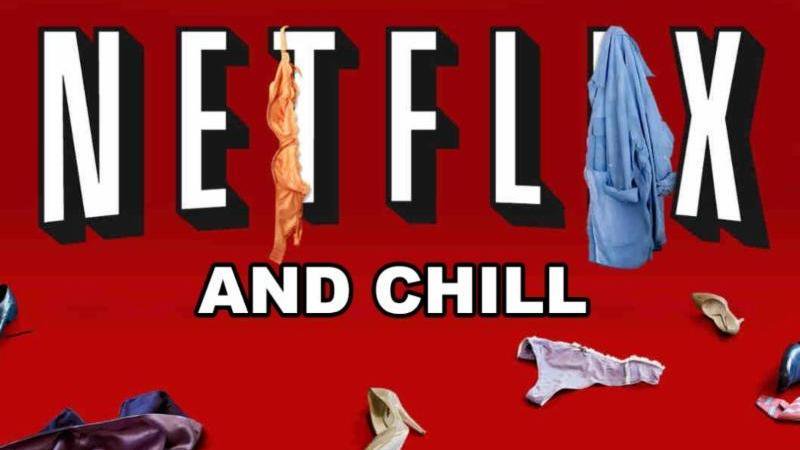}
  \caption{Example meme used in the annotation prompt: “Netflix and Chill.”}
  \Description{An image that visually illustrates the meme “Netflix and Chill,” used to guide annotators in understanding cultural misunderstandings.}
  \label{fig:netflix}
\end{figure}

To guide misunderstanding annotations in Stage 1, participants were given the following instruction with an example based on the “Netflix and Chill” meme (Figure~\ref{fig:netflix}):

\begin{lstlisting}
Describe how someone from another culture might misinterpret this meme if they lack the relevant cultural background (e.g., missing implied meanings or cultural references).
Example (Netflix and Chill): A non-U.S. viewer may take it literally as watching Netflix, not realizing it is a euphemism for romantic/sexual activity.
\end{lstlisting}

\section{BERT Classifier Performance on Varying Training Sizes}
\label{appendix:bert-performance}

We evaluated the reliability of our BERT-based classifier by training it on varying sizes of the 200 annotated samples. We reserved 50 samples as a held-out test set and trained the model on 10, 50, 100, and 150 examples. As shown in Table~\ref{tab:bert-performance}, classifier performance improved consistently with more training data and reached near-perfect levels with 150 samples, suggesting the sufficiency of our labeled set for quality filtering.

\begin{table}[t]
\centering
\begin{tabular}{cccc}
\toprule
\textbf{\# Training Samples} & \textbf{Accuracy} & \textbf{F1 Score} & \textbf{AUC-ROC} \\
\midrule
10                            & 0.891                   & 0.940             & 0.964            \\
50                            & 0.938                   & 0.963             & 0.978            \\
150                           & 0.969                   & 0.981             & 0.992            \\
200                           & 0.984                   & 0.991             & 1.000            \\
\bottomrule
\end{tabular}
\vspace{1em}
\caption{BERT classifier performance on held-out test set (50 examples) under varying training sizes.}
\vspace{-2em}
\label{tab:bert-performance}
\end{table}

\section{Additional Details on the Rewriting Process of Original Data}
\label{rewrite}
This appendix details the utilization of GPT-4, in several key stages of our data processing pipeline.
\begin{lstlisting}
You are a cultural analyst. Rewrite the input text into a standardized format without changing meaning.
Rules:
- Preserve all original keywords, slang, and cultural references (do not paraphrase them).
- Add minimal context only if needed for clarity.
- If U.S. cultural context is central, start with "In the US, this meme ..."; otherwise start with "This meme ...".
- For misunderstandings, keep the original concern but standardize phrasing.
Input:
Explanation (raw): [Explanation_Original]
Misunderstanding (raw, optional): [Misunderstanding_Original]
Output (follow exactly):
Explanation: [Standardized 1 sentence]
Potential Misunderstanding: [Standardized 1 sentence, start with "People might" / "Some viewers might"]
\end{lstlisting}

\section{Prompts for Meme Classification}
\label{prompt:classification}
\begin{lstlisting}
You are an expert in meme analysis. Given a meme image, classify it using the specified task and output format.
Task A (Knowledge Dependency):
Choose ONE label:
- Cultural Knowledge-Dependent: Requires U.S.-specific cultural knowledge (e.g., history, celebrities, media, norms).
- General Knowledge-Based: Understandable from universal experiences or common reasoning.
Output: Knowledge: [Cultural Knowledge-Dependent/General Knowledge-Based]

Task B (Topic Category):
Choose ONE label:
[Political Satire / Entertainment / Films and Cartoons / Pop Culture / Social Events]
Output: Topic: [Political Satire/Entertainment/Films and Cartoons/Pop Culture/Social Events]
\end{lstlisting}

\section{Example Data Instances}
\label{app:dataset_example}
To better illustrate the structure and labeling of our dataset, we present three representative examples below in Figure~\ref{fig:data_example}.
\begin{figure}[ht]
    \centering
    \includegraphics[width=0.6\linewidth]{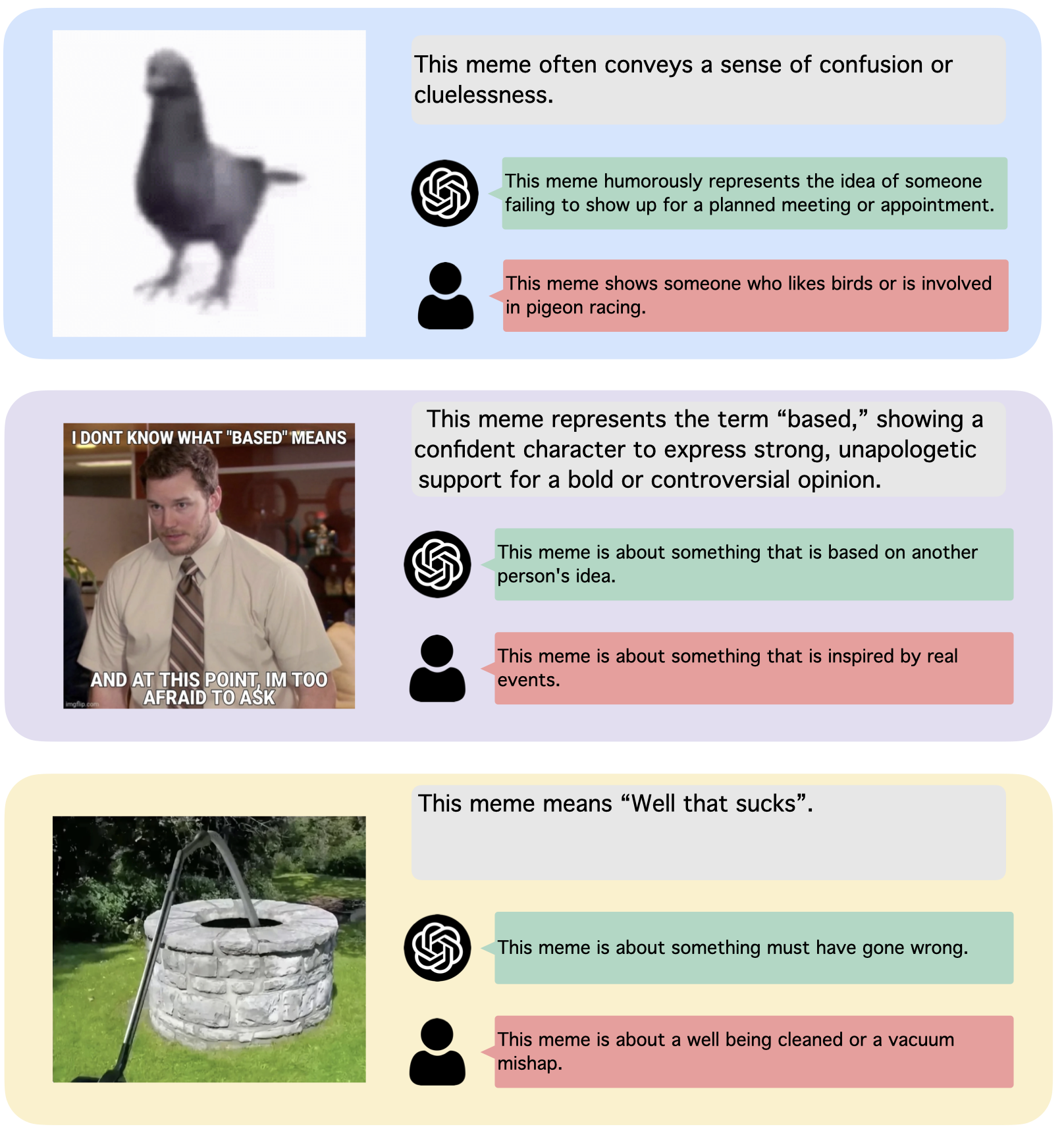}
    \Description{A data example.}
    \caption{Three example memes from the dataset. Each includes the original explanation (gray box), the GPT-4 generated interpretation (green box), and a human-annotated potential misunderstanding (red box).}
    \label{fig:data_example}
\end{figure}

\section{Multiple Choice Questions Generation}
\label{prompt:MCQ}
\begin{lstlisting}
Template 1 (Misleading Option):
Input:
- Correct answer: [Rewritten_Explanation]
- Misunderstanding: [Rewritten_Misunderstanding]
Output:
- Misleading option (20-30 words): A plausible but incorrect choice consistent with the misunderstanding.

Template 2 (Chinese-View Misunderstanding):
Instruction: Assume the role of a native Chinese viewer with limited knowledge of U.S. cultural context.
Input:
- Meme image: [Meme]
Output:
- Possible misunderstanding (20-30 words): A plausible misinterpretation based only on the image and Chinese cultural intuition.
\end{lstlisting}

\section{Prompt for Model Testing}
\label{prompt:model-test}
This appendix the structured prompts developed to evaluate how language models interpret memes across cultural and academic frameworks. 
\begin{lstlisting}
You will analyze an American meme from a specified perspective.

Perspective: [Neutral Academic / American / Chinese]

Follow the output format exactly:

Explanation: (20-30 words) Provide the meme's meaning and cultural/contextual relevance as appropriate for the selected perspective.
Misunderstanding: (20-30 words) [Include ONLY if Perspective is Neutral Academic or American] Describe a plausible misinterpretation by non-US/non-American audiences due to cultural differences.
Sentiment: [Positive/Negative/Neutral]
Emotions: [Sarcastic, Humorous, Motivational, Offensive]
\end{lstlisting}

\section{Performance Score Calculation}
\label{app:ps}
To construct PS, we aligned the expected scores of the classification tasks with the similarity metrics by setting $\mathbb{E}[\text{PS}_{\text{MCQ}}] = \mathbb{E}[\text{PS}_{\text{Sent}}] = \mathbb{E}[\text{PS}_{\text{Emo}}] = 1$. For multiple-choice and sentiment classification, where each question has three options, the expected accuracy is approximately $0.33$. For emotion labeling, where a prediction is marked correct if the ground-truth labels are a subset of the predicted labels, we calculate expected accuracy as:
\begin{equation}
\label{emo_expected}
\mathbb{E}_{\text{Emo}}[\text{Acc}] = \left(\sum_{m \in \mathcal{M}}\frac{\text{PCLS}_m}{\text{PLS}}\right) \bigg/ |\mathcal{M}|,
\end{equation}
where $\mathcal{M}$ is the set of memes, PLS denotes the number of possible label sets for each meme, and PCLS is the number of those considered correct.

The number of possible label sets (PLS) for each meme is given by: 
\begin{equation}
    \text{PLS} = \sum_{l=1}^{n}\binom{n}{l} = \sum_{l=1}^{n}\frac{n!}{l!(n - l)!},
\end{equation}
where $n$ is the number of possible labels (in this case $n = 4$), and $l$ is the number of labels chosen by the model ($l \in [1, 4]$). The number of possible correct label sets (PCLS) is then defined as:
\begin{align}
    \text{PCLS}_m &= \sum_{l_m^i=0}^{n-c_m}\binom{n-c_m}{l_m^i} \nonumber \\
    &= \sum_{l_m^i=0}^{n-c_m}\frac{(n-c_m)!}{l_m^i![(n-c_m)-l_m^i]!},
\end{align}
where $c_m$ is the number of true labels for a given meme ($c_m \in [1,4]$), and $l_m^i$ represents the number of false labels for the current meme ($l_m^i \in [0, 3]$). The expected accuracy for emotion classification is then given by Equation~\ref{emo_expected}, where $\mathcal{M}$ represents the set of all memes, with $|\mathcal{M}| = 621$. Based on our dataset, 440 memes have one emotion label, 174 have two emotion labels, 7 have three emotion labels, and no meme has all four emotion labels. Therefore, we got $\mathbb{E}_{\text{Emo}}[\text{Acc}] \approx 0.12$. To assign scores, we solve the following system of equations: 
\begin{equation}
\begin{cases}
\mathbb{E}_{\text{Emo}}[\text{Acc}] \cdot x = \mathbb{E}_{\text{MCQ}}[\text{Acc}] \cdot y, \\
x + 2y = 3\mathbb{E}[\text{PS}],
\end{cases}
\end{equation}
where $x$ is the maximum possible possible score assigned to the emotion labeling, and $y$ is the maximum possible possible score assigned to both the multiple choice question selection and sentiment labeling. Then, the final performance score can be computed as: 
\begin{align}
    \text{PS} &= \text{Sim}_\text{original} + \text{Sim}_\text{formatted} \nonumber \\
    &\quad + \text{PS}_\text{MCQ} + \text{PS}_\text{Sent} + \text{PS}_\text{Emo} \nonumber \\
    &= \text{Sim}_\text{original} + \text{Sim}_\text{formatted} \nonumber \\
    &\quad + \text{Acc}_\text{MCQ} \cdot y + \text{Acc}_\text{Sent} \cdot y + \text{Acc}_\text{Emo} \cdot x.
\end{align}

\section{Emotion Label Count Distribution}
\label{appendix:emotion-label-distribution}
For emotion prediction, we mark a response correct if the ground-truth emotion label(s) are a subset of the predicted labels. This raises a potential concern that a model could inflate accuracy by selecting all labels.

In our study, the grading rule was not disclosed to either models or human participants, making deliberate label inflation unlikely. To verify this empirically, we compare the distribution of the number of emotion labels selected per response across models and humans, summarized in Table \ref{tab:emotion_label_distribution}.

\begin{table}[t]
\centering
\begin{tabular}{lccccc}
\toprule
\textbf{\# Labels} & \textbf{Qwen} & \textbf{GPT} & \textbf{GLM} & \textbf{LLaMA} & \textbf{Human} \\
\midrule
1 & 68.9\% & 75.4\% & 52.8\% & 55.2\% & 74.5\% \\
2 & 22.9\% & 15.1\% & 32.8\% & 26.9\% & 23.8\% \\
3 & 8.2\%  & 9.5\%  & 14.4\% & 17.9\% & 1.7\%  \\
4 & 0.0\%  & 0.0\%  & 0.0\%  & 0.0\%  & 0.0\%  \\
\midrule
$\mathbb{E}[\text{\# Labels}]$ & \textbf{1.39} & \textbf{1.34} & \textbf{1.62} & \textbf{1.63} & \textbf{1.27} \\
\bottomrule
\end{tabular}
\vspace{1em}
\caption{Distribution of number of emotion labels selected per response.}
\vspace{-2em}
\label{tab:emotion_label_distribution}
\end{table}

These results show that both models and humans almost always choose 1--2 labels, occasionally 3, and never all 4. The expected number of selected labels ranges from 1.27 (human) to 1.63 (LLaMA), well below the degenerate maximum of 4, indicating no label inflation in practice.

\section{Fine-tuned Models Performance}
\label{app:ft}
\begin{figure}[t]
    \centering
    \Description{FT models performance.}
    \begin{subfigure}[b]{0.49\columnwidth}
        \includegraphics[width=\textwidth]{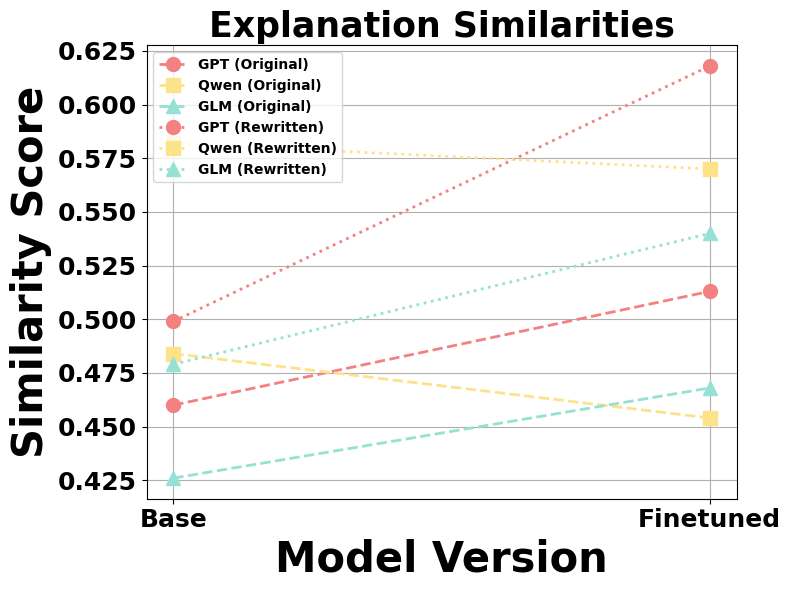}
        \caption{Explanation Similarities}
        \label{fig:sim}
    \end{subfigure}
    \hfill
    \begin{subfigure}[b]{0.49\columnwidth}
        \includegraphics[width=\textwidth]{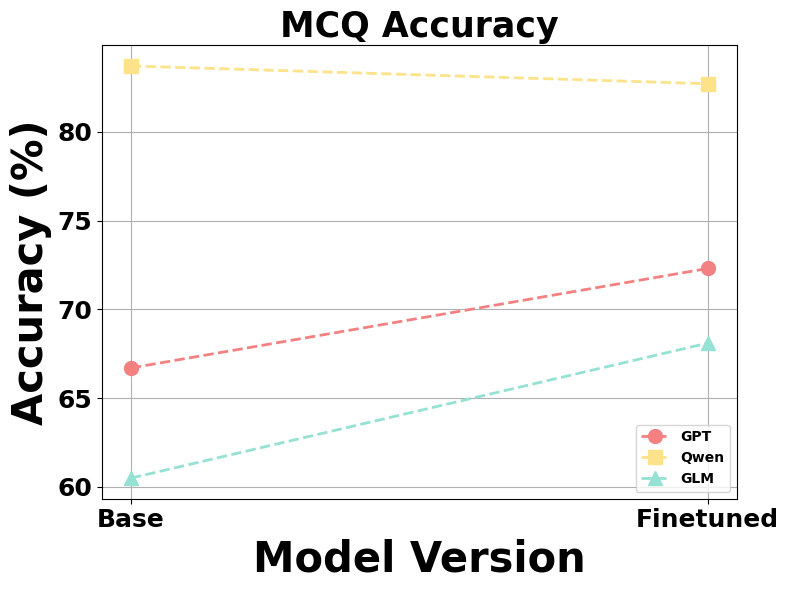}
        \caption{MCQ Accuracy}
        \label{fig:mcq}
    \end{subfigure}

    \vspace{0.5em}

    \begin{subfigure}[b]{0.49\columnwidth}
        \includegraphics[width=\textwidth]{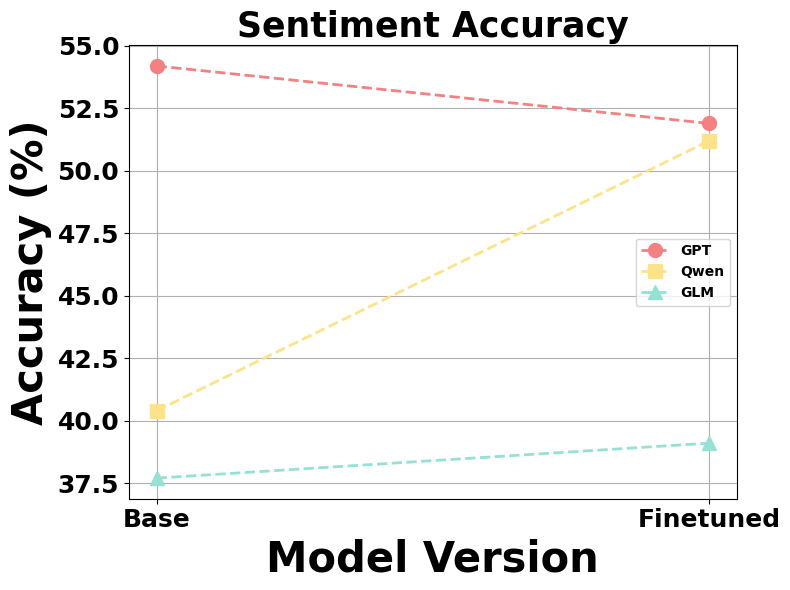}
        \caption{Sentiment Accuracy}
        \label{fig:sent}
    \end{subfigure}
    \hfill
    \begin{subfigure}[b]{0.49\columnwidth}
        \includegraphics[width=\textwidth]{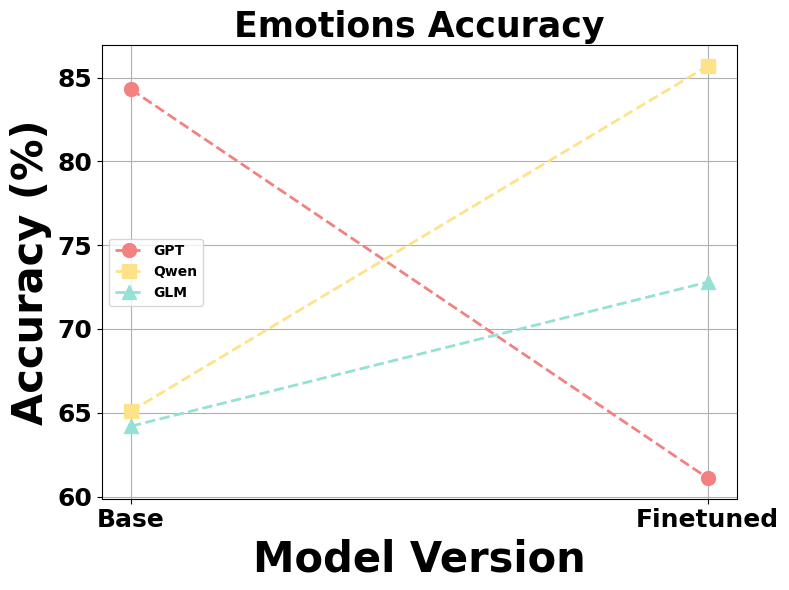}
        \caption{Emotion Accuracy}
        \label{fig:emo}
    \end{subfigure}

    \caption{Performance comparison between base models and their fine-tuned versions across different metrics.}
    \label{fig:ft_metrics}
\end{figure}

According to the results shown in Figure~\ref{fig:ft_metrics}, our dataset has the potential to enhance LLMs' ability to interpret memes, provided that overfitting does not occur. To mitigate the risk of overfitting, we recommend following our dataset curation pipeline to ensure the creation of a sufficiently large dataset.

%% file: citations.bib
@article{Grundlingh15032018,
  title={Memes as speech acts},
  author={Grundlingh, L},
  journal={Social Semiotics},
  volume={28},
  number={2},
  pages={147--168},
  year={2018}
}

@article{Mukhtar2024Memes,
  title={Memes In The Digital Age: A Sociolinguistic Examination Of Cultural Expressions And Communicative Practices Across Border},
  author={Mukhtar, Shahira and Ayyaz, Qurat Ul Ain and Khan, Sadaf and Bhopali, Atiya Muhammad Nawaz and Sajid, Muhammad Khalid Mehmood and Babbar, Allah Wasaya and others},
  journal={Educational Administration: Theory and Practice},
  volume={30},
  number={6},
  pages={1443--1455},
  year={2024}
}

@article{openai2024gpt4technicalreport,
  title={Gpt-4 technical report},
  author={Achiam, Josh and Adler, Steven and Agarwal, Sandhini and Ahmad, Lama and Akkaya, Ilge and Aleman, Florencia Leoni and Almeida, Diogo and Altenschmidt, Janko and Altman, Sam and Anadkat, Shyamal and others},
  journal={arXiv preprint arXiv:2303.08774},
  year={2023}
}

@article{Tsai02012017,
  title={Consumer engagement with brands on social network sites: A cross-cultural comparison of China and the USA},
  author={Tsai, Wan-Hsiu Sunny and Men, Linjuan Rita},
  journal={Journal of Marketing Communications},
  volume={23},
  number={1},
  pages={2--21},
  year={2017}
}

@article{bai2023qwenvlversatilevisionlanguagemodel,
  title={Qwen-vl: A frontier large vision-language model with versatile abilities},
  author={Bai, Jinze and Bai, Shuai and Yang, Shusheng and Wang, Shijie and Tan, Sinan and Wang, Peng and Lin, Junyang and Zhou, Chang and Zhou, Jingren},
  journal={arXiv preprint arXiv:2308.12966},
  year={2023}
}

@article{glm2024chatglmfamilylargelanguage,
  title={Chatglm: A family of large language models from glm-130b to glm-4 all tools},
  author={Team, GLM and Zeng, Aohan and Xu, Bin and Wang, Bowen and Zhang, Chenhui and Yin, Da and Zhang, Dan and Rojas, Diego and Feng, Guanyu and Zhao, Hanlin and others},
  journal={arXiv preprint arXiv:2406.12793},
  year={2024}
}

@inproceedings{Zhong_2024_CVPR,
  title={Multimodal Understanding of Memes with Fair Explanations},
  author={Zhong, Yang and Baghel, Bhiman Kumar},
  booktitle={CVPR},
  pages={2007--2017},
  year={2024}
}

@inproceedings{zannettou2018origins,
  title={On the origins of memes by means of fringe web communities},
  author={Zannettou, Savvas and Caulfield, Tristan and Blackburn, Jeremy and De Cristofaro, Emiliano and Sirivianos, Michael and Stringhini, Gianluca and Suarez-Tangil, Guillermo},
  booktitle={IMC},
  pages={188--202},
  year={2018}
}

@inproceedings{ramezani2023knowledge,
  title={Knowledge of cultural moral norms in large language models},
  author={Ramezani, Aida and Xu, Yang},
  booktitle={ACL},
  pages={428--446},
  year={2023}
}

@inproceedings{yao2024benchmarking,
  title={Benchmarking Machine Translation with Cultural Awareness},
  author={Yao, Binwei and Jiang, Ming and Bobinac, Tara and Yang, Diyi and Hu, Junjie},
  booktitle={Findings of EMNLP},
  pages={13078--13096},
  year={2024}
}

@inproceedings{agarwal2024mememqa,
  title={MemeMQA: Multimodal Question Answering for Memes via Rationale-Based Inferencing},
  author={Agarwal, Siddhant and Sharma, Shivam and Nakov, Preslav and Chakraborty, Tanmoy},
  booktitle={Findings of ACL},
  pages={5042--5078},
  year={2024}
}

@inproceedings{nandy2024yesbut,
  title={YesBut: A High-Quality Annotated Multimodal Dataset for evaluating Satire Comprehension capability of Vision-Language Models},
  author={Nandy, Abhilash and Agarwal, Yash and Patwa, Ashish and Das, Millon and Bansal, Aman and Raj, Ankit and Goyal, Pawan and Ganguly, Niloy},
  booktitle={EMNLP},
  pages={16878--16895},
  year={2024}
}

@inproceedings{akhtarmemex,
  title={MEMEX: Detecting Explanatory Evidence for Memes via Knowledge-Enriched Contextualization},
  author={Sharma, Shivam and Ramaneswaran, S and Arora, Udit and Akhtar, Md Shad and Chakraborty, Tanmoy},
  booktitle={ACL},
  pages={5272–-5290},
  year={2023}
}

@inproceedings{liu2022figmemes,
  title={FigMemes: A dataset for figurative language identification in politically-opinionated memes},
  author={Liu, Chen and Geigle, Gregor and Krebs, Robin and Gurevych, Iryna},
  booktitle={EMNLP},
  pages={7069--7086},
  year={2022}
}

@inproceedings{zhou2023social,
  title={Social Meme-ing: Measuring Linguistic Variation in Memes},
  author={Zhou, Naitian and Jurgens, David and Bamman, David},
  booktitle={NAACL-HLT},
  pages={3005--3024},
  year={2024}
}

@inproceedings{hwang2023memecap,
  title={MemeCap: A Dataset for Captioning and Interpreting Memes},
  author={Hwang, EunJeong and Shwartz, Vered},
  booktitle={EMNLP},
  pages={1433--1445},
  year={2023}
}

@inproceedings{nguyen2023extracting,
  title={Extracting cultural commonsense knowledge at scale},
  author={Nguyen, Tuan-Phong and Razniewski, Simon and Varde, Aparna and Weikum, Gerhard},
  booktitle={WWW},
  pages={1907--1917},
  year={2023}
}

@inproceedings{li2024culture,
  title={Culture-gen: Revealing global cultural perception in language models through natural language prompting},
  author={Li, Huihan and Jiang, Liwei and Hwang, Jena D and Kim, Hyunwoo and Santy, Sebastin and Sorensen, Taylor and Lin, Bill Yuchen and Dziri, Nouha and Ren, Xiang and Choi, Yejin},
  booktitle={COLM},
  year={2024}
}

@inproceedings{khanuja2024image,
  title={An image speaks a thousand words, but can everyone listen? On image transcreation for cultural relevance},
  author={Khanuja, Simran and Ramamoorthy, Sathyanarayanan and Song, Yueqi and Neubig, Graham},
  booktitle={EMNLP},
  pages={10258--10279},
  year={2024}
}

@article{pawar2024survey,
  title={Survey of cultural awareness in language models: Text and beyond},
  author={Pawar, Siddhesh and Park, Junyeong and Jin, Jiho and Arora, Arnav and Myung, Junho and Yadav, Srishti and Haznitrama, Faiz Ghifari and Song, Inhwa and Oh, Alice and Augenstein, Isabelle},
  journal={Computational Linguistics},
  pages={1--96},
  year={2025}
}

@inproceedings{zhao2024worldvaluesbench,
  title={WorldValuesBench: A Large-Scale Benchmark Dataset for Multi-Cultural Value Awareness of Language Models},
  author={Zhao, Wenlong and Mondal, Debanjan and Tandon, Niket and Dillion, Danica and Gray, Kurt and Gu, Yuling},
  booktitle={LREC-COLING},
  pages={17696--17706},
  year={2024}
}

@article{bui2024multi3hate,
  title={Multi3Hate: Multimodal, Multilingual, and Multicultural Hate Speech Detection with Vision-Language Models},
  author={Bui, Minh Duc and von der Wense, Katharina and Lauscher, Anne},
  journal={arXiv preprint arXiv:2411.03888},
  year={2024}
}

@inproceedings{shi2024culturebank,
  title={Culturebank: An online community-driven knowledge base towards culturally aware language technologies},
  author={Shi, Weiyan and Li, Ryan and Zhang, Yutong and Ziems, Caleb and Horesh, Raya and de Paula, Rog{\'e}rio Abreu and Yang, Diyi and others},
  booktitle={Findings of EMNLP},
  pages={4996--5025},
  year={2024}
}

@inproceedings{yin-etal-2022-geomlama,
  title={GeoMLAMA: Geo-Diverse Commonsense Probing on Multilingual Pre-Trained Language Models},
  author={Yin, Da and Bansal, Hritik and Monajatipoor, Masoud and Li, Liunian Harold and Chang, Kai-Wei},
  booktitle={EMNLP},
  pages={2039--2055},
  year={2022}
}

@inproceedings{devlin2019bertpretrainingdeepbidirectional,
  title={BERT: Pre-training of Deep Bidirectional Transformers for Language Understanding},
  author={Devlin, Jacob and Chang, Ming-Wei and Lee, Kenton and Toutanova, Kristina},
  booktitle={NAACL-HLT},
  pages={4171--4186},
  year={2019}
}

@inproceedings{sharma-etal-2020-semeval,
  title={SemEval-2020 Task 8: Memotion Analysis-the Visuo-Lingual Metaphor!},
  author={Sharma, Chhavi and Bhageria, Deepesh and Scott, William and Pykl, Srinivas and Das, Amitava and Chakraborty, Tanmoy and Pulabaigari, Viswanath and Gamb{\"a}ck, Bj{\"o}rn},
  booktitle={Proceedings of the Fourteenth Workshop on Semantic Evaluation},
  pages={759--773},
  year={2020}
}

@article{richards1987type,
  title={Type/token ratios: What do they really tell us?},
  author={Richards, Brian},
  journal={Journal of Child Language},
  volume={14},
  number={2},
  pages={201--209},
  year={1987}
}

@article{dubey2024llama,
  title={The llama 3 herd of models},
  author={Dubey, Abhimanyu and Jauhri, Abhinav and Pandey, Abhinav and Kadian, Abhishek and Al-Dahle, Ahmad and Letman, Aiesha and Mathur, Akhil and Schelten, Alan and Yang, Amy and Fan, Angela and others},
  journal={arXiv preprint arXiv:2407.21783},
  year={2024}
}

@article{hurst2024gpt,
  title={Gpt-4o system card},
  author={Hurst, Aaron and Lerer, Adam and Goucher, Adam P and Perelman, Adam and Ramesh, Aditya and Clark, Aidan and Ostrow, AJ and Welihinda, Akila and Hayes, Alan and Radford, Alec and others},
  journal={arXiv preprint arXiv:2410.21276},
  year={2024}
}

@inproceedings{liu2021visually,
  title={Visually Grounded Reasoning across Languages and Cultures},
  author={Liu, Fangyu and Bugliarello, Emanuele and Ponti, Edoardo Maria and Reddy, Siva and Collier, Nigel and Elliott, Desmond},
  booktitle={EMNLP},
  pages={10467--10485},
  year={2021}
}

@inproceedings{li2023cultural,
  title={Cultural concept adaptation on multimodal reasoning},
  author={Li, Zhi and Zhang, Yin},
  booktitle={EMNLP},
  pages={262--276},
  year={2023}
}

@inproceedings{huang2023culturally,
  title={Culturally aware natural language inference},
  author={Huang, Jing and Yang, Diyi},
  booktitle={Findings of EMNLP},
  pages={7591--7609},
  year={2023}
}

@inproceedings{kabra2023multi,
  title={Multi-lingual and Multi-cultural Figurative Language Understanding},
  author={Kabra, Anubha and Liu, Emmy and Khanuja, Simran and Aji, Alham Fikri and Winata, Genta and Cahyawijaya, Samuel and Aremu, Anuoluwapo and Ogayo, Perez and Neubig, Graham},
  booktitle={Findings of ACL},
  pages={8269--8284},
  year={2023}
}

@inproceedings{10596638,
  title={A Comprehensive Review of Sentiment Analysis: Techniques, Datasets, Limitations, and Future Scope},
  author={Rani, Neetu and Walia, Ranjan and others},
  booktitle={International Conference on Computational Intelligence and Communication Technologies},
  pages={403--409},
  year={2024}
}

@misc{gu2025surveyllmasajudge,
      title={A Survey on LLM-as-a-Judge}, 
      author={Jiawei Gu and Xuhui Jiang and Zhichao Shi and Hexiang Tan and Xuehao Zhai and Chengjin Xu and Wei Li and Yinghan Shen and Shengjie Ma and Honghao Liu and Saizhuo Wang and Kun Zhang and Yuanzhuo Wang and Wen Gao and Lionel Ni and Jian Guo},
      year={2025},
      eprint={2411.15594},
      archivePrefix={arXiv},
      primaryClass={cs.CL},
      url={https://arxiv.org/abs/2411.15594}, 
}
